\documentclass[lettersize,journal]{IEEEtran}
\usepackage{amsmath,amsfonts}
\usepackage{algorithmic}
\usepackage{algorithm}
\usepackage{array}
\usepackage[caption=false,font=normalsize,labelfont=sf,textfont=sf]{subfig}
\usepackage{textcomp}
\usepackage{stfloats}
\usepackage{url}
\usepackage{verbatim}
\usepackage{graphicx}
\usepackage{cite}
\usepackage{booktabs}
\usepackage{float}
\usepackage{multicol}
\usepackage{hyperref}
\usepackage{csquotes}
\usepackage{longtable} 
\hypersetup{
    colorlinks=true,
    linkcolor=blue,
    filecolor=magenta,      
    urlcolor=cyan,
    pdftitle={Overleaf Example},
    pdfpagemode=FullScreen,
    }

\begin{document}

\title{ID-Card Synthetic Generation: Toward a Simulated Bona fide Dataset}

\author{
    Qingwen Zeng,
    Juan~E.~Tapia,~\IEEEmembership{Senior~Member,~IEEE,}
    Izan Garcia,
    Juan M.~Espín,
    and~Christoph~Busch,~\IEEEmembership{Senior~Member,~IEEE}%
\thanks{Juan E. Tapia and Christoph Busch are with the da/sec-Biometrics and Internet Security Research Group, Hochschule Darmstadt, Germany, e-mail: \{\href{juan.tapia-farias@h-da.de}{juan.tapia-farias}, \href{christoph.busch@h-da.de}{christoph.busch}\}@h-da.de.}%
\thanks{Qingwen Zeng is with the DTU, Denmark, email: \href{s232892@student.dtu.dk}{s232892@student.dtu.dk}.}%
\thanks{Izan Garcia and Juan M. Espín are with Facephi company, R\&D department, Alicante, Spain, email: \href{jmespin@facephi.com}{izangarcia, jmespin{@facephi.com}}.
}%
\thanks{Corresponding author: Juan E Tapia. Manuscript received Month DD, YYYY; revised Month DD, YYYY.}}


\markboth{Journal of \LaTeX\ Class Files,~Vol.~14, No.~8, August~2021}%
{Shell \MakeLowercase{\textit{et al.}}: A Sample Article Using IEEEtran.cls for IEEE Journals}


\maketitle

\begin{abstract}
Nowadays, the development of a Presentation Attack Detection (PAD) system for ID cards presents a challenge due to the lack of images available to train a robust PAD system and the increase in diversity of possible attack instrument species. Today, most algorithms focus on generating attack samples and do not take into account the limited number of bona fide images. This work is one of the first to propose a method for mimicking bona fide images by generating synthetic versions of them using Stable Diffusion, which may help improve the generalisation capabilities of the detector. Furthermore, the new images generated are evaluated in a system trained from scratch and in a commercial solution. The PAD system yields an interesting result, as it identifies our images as bona fide, which has a positive impact on detection performance and data restrictions.
\end{abstract}


\section{Introduction}
\IEEEPARstart{T}he generation of synthetic images for training of Presentation Attack Detection (PAD) has increased in recent years, motivated by the currently lower generalisation capabilities of the PAD ID card systems to classify other unseen ID documents. Currently, synthetic images cannot be used in the training process to improve the system because they are focused only on attack scenarios and do not follow the ICAO standard\footnote{\url{https://www.icao.int/sites/default/files/publications/DocSeries/9303_p1_cons_en.pdf}}. The lack of bona fide ID card images available for training a robust PAD system is a real challenge.

To address these limitations, most databases focus on creating attack images, which involves printing a digital version on a polyvinyl chloride (PVC) template ID card as a bona fide ID card and then using this physical printed version to generate new attacks from those IDs. These templates are later printed on glossy paper to produce printed attacks. The same image is displayed on different screens to obtain the screen attack, respectively, and train a PAD system. 

Another option explored is to create synthetic images using generative adversarial models (GANs), which currently present low-fidelity reproductions of the original properties in terms of noise reduction and similarity with bona fide samples. Furthermore, most of them do not adhere to the ICAO standard for creating realistic samples and can therefore not be used as an alternative to improving the PAD system performance.

Recent research in biometrics has been focused mainly on generating synthetic face images for training of face PAD systems. Image generation has reached a remarkable level of realism, with many algorithms contributing to this success~\cite{Karras2019stylegan2, stylegan3}. On the other hand, generating ID documents is very challenging, intricate, and more difficult than face image generation because it involves faces in neutral expressions, ICAO-compliant, dynamic text, signatures, and a Machine-Readable Zone (MRZ) that contains checksum digits or letters. The current state-of-the-art model generates ID documents with low similarity to bona fide images, which often contain a lot of noise, artefacts, blurring, and illegible text. 

Motivated by the previous challenges, we proposed and explored two generative methods. The first is based on Stable Diffusion to generate hybrid ID card images, which means using computer vision techniques to generate synthetic face images on the one hand and to create text and signatures on the other hand. The second one explored is a one-shot generation approach, which involves synthetically generating all the ID card information based on Stable Diffusion. We aim to enhance the PAD ID card system by incorporating \textit{simulated bona fide}\footnote{A simulated bona fide is a synthetic image.} ID card images that replace or complement bona fide images.

The main contributions of this work are:

\begin{itemize}
    \item A generation of face image ICAO compliance has been explored to create images focusing on ID cards.
    \item A hybrid method for generating \textit{simulated bona fide} ID card images similar to bona fide distribution.
    \item A second method for generating \textit{simulated bona fide} ID card images based on a one-shot approach to generate images similar to bona fide distribution.
    \item We demonstrated the pros and cons of these two approaches while evaluating a new PAD approach and a commercial off-the-shelf (COTS) system.
    \item This work depicted the relevance of bona fide images and highlighted the constraints in generating new ID cards.
    \item The \textit{simulated bona fide} ID card images generated will be available for research purposes only.
\end{itemize}

The rest of the paper is organised as follows: Section~\ref{sec:related} reviews the related works. Section~\ref{sec:metrics} explains the metrics used to analyse the experiments. Section~\ref{sec:methods} describes the techniques used. Section~\ref{sec:data} shows and describes all the datasets used in this work. Section~\ref{sec:results} explains the experiments and results, and Section~\ref{sec:conclusion} draws the findings of this work.

\section{Related work}
\label{sec:related}

\subsection{PAD on ID cards}

One of the main challenges in the state of the art is the lack of ID card document images because of privacy concerns and constraints. Today there is a limited number of open datasets with ID card images are available, such as MIDV500~\cite{MIDV500},~MID-Holo~\cite{MIDV2020}, DLC2021~\cite{MIDVDLC2021},~KID34~\cite{KID34K},~ID-Net~\cite{IDNet}, and others.

These datasets contained fewer subjects and hundreds of thousands of images generated for \textit{simulated synthetic} attacks. As a result, the system exhibited strong overfitting, reduced generalisation capabilities, and failed to adhere to the ICAO rules for representing realistic ID cards. Currently, methods based on private datasets have achieved better generalisation capabilities due to the quality and quantity of bona fide images and the number of samples per image available \cite{GONZALEZ-PR}.

Benalcazar et al.~\cite{benalcazar} proposed a method to generate synthetic images based on a generative adversarial network. This work showed promising results and was one of the first to raise the relevance of developing new algorithms to train PAD on ID Card systems. 

Markan et al.~\cite{Markham} proposed a method based on Style transfer algorithms, such as CycleGANs\cite{isola2017image} and Pix2pixHD, in open-set datasets to generate synthetic attacks for ID cards, such as printed and screen. 

Park et al.~\cite{KID34K} proposed a new dataset based on Chinese ID cards, focusing on printed and screen attacks. However, this dataset did not adhere to the ICAO standards, rendering the \textit{simulated bona fide} examples unrealistic for a real PAD system. This kind of dataset can be used to test the system as a basic threat instead of helping to train a more robust classifier.

The IDNet dataset~\cite{IDNet} contains 837,060 images of synthetically generated identity documents, totalling approximately 490 gigabytes. These images are categorised into 20 types, which include driver's licenses from 10 U.S. states and passports and ID cards from 10 European countries. However, private datasets with more subjects, bona fide documents, attacks, and images per subject have reported more reliable results.

Tapia et al.~\cite{Tapia-IJCB2024} organised the first and second PAD competitions on ID cards in order to create a baseline for the PAD system in challenging scenarios. The competition results showed that detecting PAD attacks in a cross-dataset scenario remains challenging and that current technology can not achieve generalisation towards other countries. 

Cheng et al.~\cite{Chen-tifs} proposed a method to generate specific attack images in order to learn print and moire patterns available in the screen attacks.

A summary of the most relevant private and open-set datasets available is summarised in Table~\ref{tab:survey_Idcard}.

\subsection{Face image generation}

Generative Adversarial Networks (GANs) have been explored for generating high-quality face images in realistic scenarios, achieving remarkable realism~\cite{Karras2019stylegan2,stylegan3}. The GANs have been applied to several scenarios, tasks and algorithms. These algorithms help improve the training process of many biometric algorithms, complementing the lack of images due to privacy concerns.

StyleGAN algorithm, based on GANs, focuses on style transformation and provides more control over data generation. This model is primarily known for its high-resolution and lifelike rendering of faces. It has been extended to new versions, such as StyleGANV2~\cite{Karras2019stylegan2} and StyleGANV3~\cite{stylegan3}. The new StyleGANV3 generator achieved the performance of StyleGANV2 in terms of quality degradation, as measured with the Frechet Inception Distance metric, while being slightly more computationally intensive. However, it cannot handle the details associated with small movements, such as hair, very well.

Diffusion models~\cite{SD3} have been trained to generate data from noise. They are trained to reverse the process of transforming data into random noise. By utilising the approximation and generalisation capabilities of convolutional neural networks, these models can create new data points that are not found in the training dataset while still adhering to the distribution of that training data. The Stable Diffusion-V3 improves existing noise sampling techniques for training rectified flow models by biasing them towards perceptually relevant scales.

Flux-1-dev~\cite{bitflux,gao2024eraseanything} is an advanced text-to-image model created by Black Forest Labs. It builds upon the foundation of diffusion models, such as Stable Diffusion. With a 12 billion parameter architecture, Flux-1-dev employs a combination of multimodal diffusion and parallel transformer blocks, effectively generating detailed, high-quality images from text prompts. Flux-1-dev surpasses Stable Diffusion in handling complex scenes, speed, and image quality. While Stable Diffusion excels in photorealism and fine-tuning, Flux-1-dev effectively manages intricate compositions, thanks to its advanced architecture, which features parallel attention layers and guidance distillation techniques.

The Variational Hyper-Encoding Networks (HyperVAE)~\cite{nguyen} is a powerful deep generative model that learns to generate the parameters of Variational AutoEncoder (VAE) networks for modelling the distribution of different tasks. The versatility of the HyperVAE in producing VAE models can be applied to various challenges where model flexibility is required, including density estimation, outlier detection, and novelty detection. This model provides a Bayesian Optimisation, to search in the low-dimensional encoding space of VAE. Once a low-dimensional design is suggested, we can decode it to the corresponding high-dimensional design.

\begin{table*}[h]
\centering
\scriptsize
\caption{Nº Img represents the total number of documents. Nº users (number of different documents). Generated from a template represents a description of the nature of the ID-template or bona fide images. ICAO represents whether the dataset follows the standard.
IDC: ID card, DLI: Driver's license, PSP: Passport, FOC: Resident permit}
\label{tab:survey_Idcard}
\begin{tabular}{|c|c|c|c|c|c|c|}
\hline
Database &
  \begin{tabular}[c]{@{}c@{}}Nª\\ Images\end{tabular} &
  \begin{tabular}[c]{@{}c@{}}Nº\\ Users\end{tabular} &
  \begin{tabular}[c]{@{}c@{}}Type of\\ Attacks\end{tabular} &
  \begin{tabular}[c]{@{}c@{}}Type of\\ Document\end{tabular} &
  \begin{tabular}[c]{@{}c@{}}Generate or\\ Capture \\ from template\end{tabular} &
  \begin{tabular}[c]{@{}c@{}}ICAO\\ Compliant\end{tabular} \\ \hline
\begin{tabular}[c]{@{}c@{}}KID34K \cite{KID34K}\end{tabular} &
  $\sim$35K &
  82 &
  \begin{tabular}[c]{@{}c@{}}PVC, \\ Screen,\end{tabular} &
  \begin{tabular}[c]{@{}c@{}}IDC,\\ DLI\end{tabular} &
  Yes &
  No \\ \hline
\begin{tabular}[c]{@{}c@{}}MIDV\_500 \cite{MIDV500}\end{tabular} &
  $\sim$10K &
  50 &
  PVC &
  \begin{tabular}[c]{@{}c@{}}IDC,\\ DLI, \\ PSP, \\ FOC\end{tabular} &
  Yes &
  Partially \\ \hline
\begin{tabular}[c]{@{}c@{}}MIDV2019 \cite{MIDV2019}\end{tabular} &
  $\sim$6K &
  50 &
  PVC &
  \begin{tabular}[c]{@{}c@{}}IDC,\\ DLI,\\ PSP,\\ FOC\end{tabular} &
  Yes &
  Partially \\ \hline
\begin{tabular}[c]{@{}c@{}}MIDV2020 \cite{MIDV2020}\end{tabular} &
  $\sim$72K &
  1k &
  PVC &
  \begin{tabular}[c]{@{}c@{}}IDC,\\ PSP\end{tabular} &
  Yes &
  Partially \\ \hline
\begin{tabular}[c]{@{}c@{}}MIDVDLC-2021 \cite{MIDVDLC2021}\end{tabular} &
  $\sim$79K &
  80 &
  \begin{tabular}[c]{@{}c@{}}PVC, \\ Screen, \\ Print,\end{tabular} &
  \begin{tabular}[c]{@{}c@{}}IDC,\\ PSP\end{tabular} &
  Yes &
  Partially \\ \hline
MIDV-Holo \cite{MIDVHolo} &
  $\sim$30K &
  100 &
  \begin{tabular}[c]{@{}c@{}}PVC, \\ Digital\\ tampering\end{tabular} &
  \begin{tabular}[c]{@{}c@{}}IDC,\\ PSP\end{tabular} &
  Yes &
  Partially \\ \hline
\begin{tabular}[c]{@{}c@{}}Synt-ID-Card \cite{benalcazar}\end{tabular} &
  $\sim$4K &
  - &
  Synthetic &
  IDC &
  No &
  Yes \\ \hline
IDNet \cite{IDNet} &
  $\sim$600K &
  $\sim$120K &
  \begin{tabular}[c]{@{}c@{}}Digital\\ tampering\end{tabular} &
  \begin{tabular}[c]{@{}c@{}}IDC,\\ PSP,\\ DLI\end{tabular} &
  Yes &
  Partially \\ \hline
SIDTD \cite{SIDTD} &
  $\sim$8K &
  1K &
  \begin{tabular}[c]{@{}c@{}}Digital\\ tampering,\end{tabular} &
  \begin{tabular}[c]{@{}c@{}}IDC,\\ PSP\end{tabular} &
  Yes &
  Partially \\ \hline
FMIDV \cite{FMIDV} &
  $\sim$28K &
  1K &
  \begin{tabular}[c]{@{}c@{}}Digital \\ tampering\end{tabular} &
  \begin{tabular}[c]{@{}c@{}}IDC,\\ PSP\end{tabular} &
  Yes &
  Partially \\ \hline
\begin{tabular}[c]{@{}c@{}}Gonzalez et al. \cite{Gonzalez-Tbiom}\end{tabular} &
  10K &
  $\sim$30K &
  \begin{tabular}[c]{@{}c@{}}Border,\\ Print,\\ Screen\\ PVC\end{tabular} &
  IDC &
  No &
  Yes \\ \hline
\begin{tabular}[c]{@{}c@{}}Gonzalez et al. \cite{GONZALEZ-PR}\end{tabular} &
  70K &
  $\sim$220K &
  \begin{tabular}[c]{@{}c@{}}Border,\\ Print,\\ Screen\\ PVC\end{tabular} &
  IDC &
  No &
  Yes \\ \hline
\end{tabular}%
\end{table*}

\section{Metric}
\label{sec:metrics}
This section describes the metrics used to evaluate image generation and PAD ID card performance.
\subsection{Sample Quality with Frechet Inception Distance}

Frechet Inception Distance (FID) compares the similarity between two groups of images, $A$ and $B$. First, to compute the FID, all images from set $A$ and set $B$ have to be processed by an InceptionV3 network, pre-trained on ImageNet~\cite{fid}. Then, the 2,048-dimensional feature vector of the Inception-V3-Pool3 Layer is stored for each image. Finally, the distributions of $A$ and $B$ in the feature space are compared using Equation~\ref{eq:fid}, where $\mu_A$ and $\mu_B$ are the mean values of the distributions $A$ and $B$, respectively, and $\Sigma_A$ and $\Sigma_B$ are the covariances of the two distributions.

\begin{equation}\label{eq:fid}
    FID = \| \mu_A - \mu_B \|^2 + Tr \left( \Sigma_A + \Sigma_B -2  ( \Sigma_A \cdot \Sigma_B ) ^{1/2} \right)
\end{equation}

\subsection{PAD metrics}

The international standard ISO/IEC 30107-3\footnote{\url{https://www.iso.org/standard/67381.html}} presents methodologies for evaluating the detection performance of PAD algorithms for biometric systems. The Attack Presentation Classification Error Rate (APCER) metric measures the proportion of attack presentations for each Presentation Attack Instrument (PAI) incorrectly classified as bona fide presentations. This metric is calculated for each PAI, considering the worst-case scenario. Equation~\ref{eq:apcer} details how to compute the APCER metric, in which the value of $N_{PAIS}$ corresponds to the number of attack presentation, where $RES_{i}$ for the $i$th image is $1$ if the algorithm classifies it as an attack presentation, or $0$ if it is classified as a bona fide presentation (real image).

\begin{equation}\label{eq:apcer}
    {APCER_{PAIS}}=1 - (\frac{1}{N_{PAIS}})\sum_{i=1}^{N_{PAIS}}RES_{i}
\end{equation}

Additionally, the Bona fide Presentation Classification Error Rate (BPCER) metric measures the proportion of bona fide presentations incorrectly classified as attack presentations. The BPCER metric is formulated according to equation~\ref{eq:bpcer}, where $N_{BF}$ corresponds to the number of bona fide presentation images, and $RES_{i}$ takes identical values to those of the APCER metric. Further on, the Equal Error Rate (EER) is also reported.
The EER is the value when the APCER equals the BPCER.

\begin{equation}\label{eq:bpcer}
    BPCER=\frac{\sum_{i=1}^{N_{BF}}RES_{i}}{N_{BF}}
\end{equation}

These metrics effectively measure the degree to which the algorithm confuses presentations of attack images with bona fide images and vice versa. The APCER and BPCER metrics depend on a decision threshold.

\section{Method}
\label{sec:methods}

This work proposes and develops two methods to generate ID card images, considering two pipelines.
The first method involves a hybrid approach that combines the synthetic generation of face images with computer vision libraries to create random demographic text related to each ID card, thereby obtaining a detailed representation of the ID card.

The second proposed a method to generate the entire ID card in one step using Stable Diffusion models. Both proposals take into account features of three algorithms: Stable Diffusion-V3, Flux-1-dev, and HyperVAE.

\subsection{Hybrid-Generation}
The hybrid or step-by-step generation method is based on the Chilean ID card, motivated by access to the set of bona fide images obtained from a private company for research purposes only. These images enable us to measure the deviation between our \textit{simulated bona fide} images and genuine images. This method involves the following steps:
ID template, component segmentation and collection, frontal-facing facial images generation-filtering, post-processing, and layer composite.


\begin{figure*}[]
\centering
\includegraphics[scale=0.87]{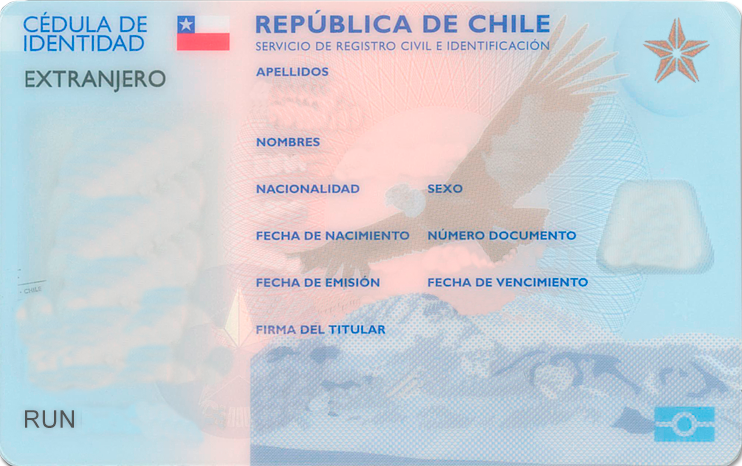}
\includegraphics[scale=0.26]{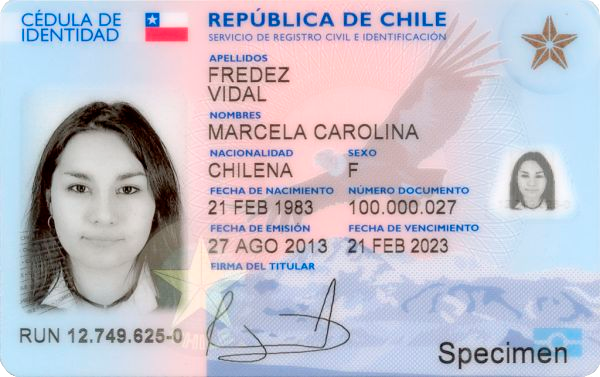}
\includegraphics[scale=0.9]{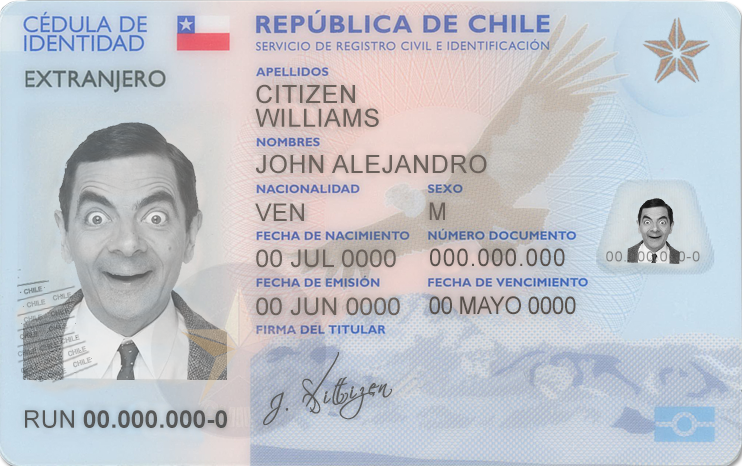}
\caption{Left to right: Empty template. Middle: The Reference of the ID without EXTRANJERO. Right: The Reference of the ID with EXTRANJERO.}\label{fig:example-CL-images}
\end{figure*}

\subsubsection{ID template} 
The ID templates used for synthesis are derived from Chile’s official identity document; an example presented in Fig. \ref{fig:example-CL-images}. The templates retains the overall structure and official elements of the document, including: the Chilean flag and the words \enquote{REPÚBLICA DE CHILE} at the top, the name of the Civil Registry and Identification Service (Servicio de Registro Civil e Identificación), the Andes Mountains pattern in the background of the document, the eagle and star anti-counterfeiting signs, and the symbol indicating the chip function in the lower right corner. 

In addition, the titles of the various fields on the document such as surname, first name, nationality, gender, date of birth, document number, issue and expiration date, and signature of the holder, but all specific personal information has been deleted, there is no detailed name, photo, signature, date of birth, nationality, document number, etc., and the RUN (Rol Único Nacional) field is also blank.

The task requires separate synthesis processes for two different templates; however, both follow a unified methodology. There are two primary distinctions between the templates. First, the presence of the word \enquote{EXTRANJERO} in the upper right corner of one Chilean identity card template signifies that it is specifically issued to non-Chilean nationals, serving as a legal residence identification card for foreigners in Chile. For clarity, this template will be referred to as the \enquote{template with EXTRANJERO.} In contrast, the absence of this \enquote{EXTRANJERO} in the other template represents that it belongs to Chilean citizens, thus referred to as the \enquote{template without EXTRANJERO.} Secondly, the templates have different resolutions: the template with EXTRANJERO has a resolution of $742 \times 466$ pixels, whereas the template without EXTRANJERO has a resolution of $600 \times 377$ pixels.\\

\subsubsection{Components segmentation and collection}
The ID templates constitute the primary inputs, while ID references provide the corresponding observations. The ID references are presented in Fig. \ref{fig:example-CL-images} and represent the idealised outputs generated by the synthesis process applied to each template. 

According to the reference, the elements of Chile’s official identity document that require substantial random generation can be broadly divided into three main categories: frontal-facing facial images, textual content, and signatures. 

Synthesising the entire ID document while ensuring consistency and accuracy across these categories presents significant challenges for generative models. Common issues include low-quality facial images, inconsistent fonts within textual content, and layout structures that deviate from the original reference.

To address these issues, each of the primary categories can be further decomposed into more detailed, well-defined components with a corresponding spatial coordinate. For example, frontal-facing facial images may comprise two distinct photographs adhering to different image parameters, while the textual content includes multiple personal information fields, each with specific formatting and size. 

\begin{figure}[H]
\centering
\includegraphics[scale=0.23]{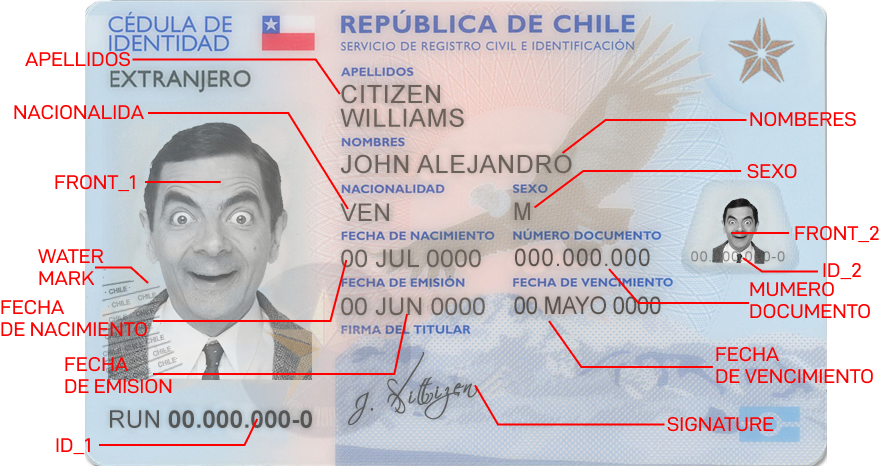}
\caption{The Components Segmentation of the Chilean ID card.}
\label{fig: Components segmentation}
\end{figure}

A comprehensive segmentation of these components is illustrated in Fig.~\ref{fig: Components segmentation}. Given this structure, we propose a modular synthesis strategy in which each component is generated independently and then accurately composited onto the ID template at predefined coordinates, i.e., Hybrid Generation. This method reduces the synthesis complexity and improves flexibility, allowing each category to be generated or collected using individually optimised and specialised methods. 
\vspace{-0.3cm}

For generating frontal-facing facial images, a generative model is employed to produce a large number of high-quality images that maintain diversity while adhering to the corresponding regulations.

For the textual component, a comparative analysis of font styles determined that \textit{Arial} most closely matches the desired appearance. 
Meanwhile, the textual content in different personal information fields exhibits distinct formatting rules. For example, the RUN (Rol Único Nacional), which is the Chilean national identification number, follows a specific structure: a randomly assigned number correlated with the Chilean population, followed by a hyphen and a check digit computed using a modulo $11$ algorithm. Each of these formatting rules should be documented and preserved to ensure consistency in the subsequent data synthesis process. 

\begin{figure}[H]
\centering
\includegraphics[scale=0.25]{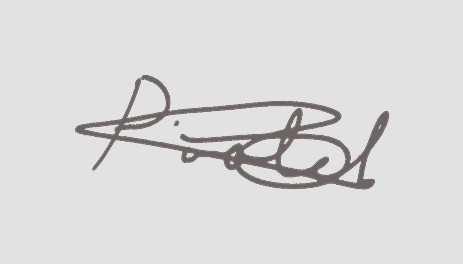}
\caption{A Signature Example from GPDS 1–150.}
\label{fig: Signature}
\end{figure}

For the signature, we adopt the GPDS 1–150, a subset from the GPDS-960 dataset \cite{vargas2007off} as the source of handwritten signature images, providing a standardised and widely accepted benchmark for signature data. A signature example is illustrated in Fig.~\ref{fig: Signature}.
\vspace{-0.3cm}

Due to the inaccessibility of the official layout from the Chilean government regarding the identity document, it is necessary to manually extract the layout parameters of each component based on the reference. These parameters include the size, coordinates, transparency, colour, and rotation angle of each component. To achieve this process, we employ Figma~\footnote{\url{https://www.figma.com/}}, a vector graphics editor, to perform precise measurements. The layout information is then stored in a JSON file format, serving as input for the subsequent synthesis process. 

\subsubsection{Frontal-facing facial images generation and filtering}
According to Exempt Resolution No. 466 issued by the Ministry of Justice and Human Rights and the Civil Registry and Identification Service of Chile \footnote{\url{https://www.bcn.cl/leychile/navegar?i=1209132}}, Chile’s
official identity document should comply with ICAO Doc 9303 requirements \footnote{\url{https://www.icao.int/publications/pages/publication.aspx?docnum=9303}}. Therefore, our objective is to generate high-quality, frontal-facing facial images that meet the ICAO Doc 9303 guidelines.

We refer to the generation process of the ONOT, an ICAO-compliant synthetic mugshot dataset~\cite{onot}. High‐quality frontal-facing facial images that comply with ICAO Doc 9303 are generated through the Stable Diffusion Web UI \footnote{\url{https://github.com/AUTOMATIC1111/stable-diffusion-webui}}. This GitHub repository provides a convenient interface for interacting with the Stable Diffusion model~\cite{rombach2022high}. 

Specifically, in the generation process, most of the parameter settings follow those of ONOT. Each front‐facing facial image has a resolution of $512\times 512$ pixels. During the sampling process, we employ the \textit{DPM++ SDE Karras} as simpler with $25$ sampling steps. However, several modifications are applied to align the generation process to the requirements of this task. 
Since the task does not require identity consistency across samples, the specific identity generation of each image is considered irrelevant to the generation process. To ensure reproducibility, we fix the random seed at $-1$.

The generation process is primarily controlled by textual prompts and pretrained model checkpoints, which collectively define our optimisation objective. Specifically, starting from the baseline generation settings of ONOT, we performed multiple rounds of prompt and checkpoint optimisations. As a result, several key prompts were added to follow the constraints of ICAO Doc 9303, for example, the inclusion of teeth in negative prompts. The details of the prompt setting are presented in the Table~\ref{tab:prompts}. 

The original checkpoint was replaced by a variant, denoted as \textit{realisticVisionV60B1\_v51HyperVAE} \footnote{\url{https://huggingface.co/JCTN/Juggernaut/blob/main/realisticVisionV60B1_v51HyperVAE.safetensors}}, to obtain more vivid colour performance. 

Following the generation of approximately 5,000 images, the subsequent step involves quality assessment to ensure compliance with quality standards. Therefore, we utilise the Open-source Face Image Quality (OFIQ)~\cite{Merkle-OFIQ-Report-240930} to filter the generated images. 

The OFIQ evaluates images based on $27$ quality measures. Each indicator is independently assessed by distinct algorithms, reporting the utility of a quality component on a standardised scale of $0$ to $100$, and complemented by a comprehensive unified quality score, which typically ranges from $0$ to $100$ where higher values indicate better utility. 

After evaluation, the generated images demonstrate strong performance across most quality measures. However, certain components exhibit score variability. To ensure robustness, we establish empirically derived thresholds for key measures, thereby guaranteeing only high-quality, frontal-facing facial images. The specified thresholds are detailed in the Table~\ref{tab: OFIQ threshold}.

\begin{table}[H]
\centering
\scriptsize
\caption{OFIQ threshold}
\label{tab: OFIQ threshold}
\begin{tabular}{|c|c|}
\hline
OFIQ Quality Measures & Thresholds \\ \hline
UnifiedQualityScore.scalar & 75 \\ \hline
BackgroundUniformity.scalar & 70 \\ \hline
IlluminationUniformity.scalar & 65 \\ \hline
LuminanceVariance.scalar & 65 \\ \hline
OverExposurePrevention.scalar & 80 \\ \hline
InterEyeDistance.scalar & 80 \\ \hline
HeadSize.scalar & 30 \\ \hline
MarginAboveOfTheFaceImage.scalar & 35 \\ \hline
\end{tabular}
\end{table}

Suppose the initial dataset fails to filter an adequate number of images that meet these quality measures, which in turn affects the synthesis capabilities, then we employ a regeneration by the Image-to-Image function within the \textit{Stable Diffusion Web UI}. The frontal-facing facial images have been selected as high-quality in the initial generation and can be treated as a template. Reserving the same parameters as used in the initial generation could significantly enhance the quality of the expanded dataset, thereby increasing the likelihood of images meeting OFIQ criteria. 

The next steps involve the filtering of the generated images against the OFIQ threshold criteria in Table~\ref{tab: OFIQ threshold}. This iterative cycle of image generation, augmentation, and filtering continues until the volume of high-quality, compliant images is sufficient to meet our requirements.

\begin{table*}[]
\centering
\caption{The negative prompt used for generating the images, the template of the positive prompt, and one example of a positive prompt.}
\label{tab:prompts}
\begin{tabular}{@{}p{3cm}p{11cm}@{}}
\toprule
\textbf{Prompt Type} & \textbf{Content} \\ \midrule
Negative prompt & { \raggedright (deformed iris, deformed pupils, semi-realistic, CGI, 3D, render, sketch, cartoon, drawing, anime:1.4), text, close-up, cropped, out of frame, worst quality, low quality, jpeg artifacts, ugly, duplicate, morbid, mutilated, extra fingers, mutated hands, poorly drawn hands, poorly drawn face, mutation, deformed, blurry, dehydrated, bad anatomy, bad proportions, extra limbs, cloned face, disfigured, gross proportions, malformed limbs, missing arms, missing legs, extra arms, extra legs, fused fingers, too many fingers, long neck, hair in front of the eyes, hat, (shadows), (three-quarter pose), (face in profile:1.1), teeth, full body, side view. \par} \\[10pt]
Positive Prompt Template & { \raggedright RAW front photo, face portrait photo of (\{years\} years old:1.1), \{ethnicity\} (\{gender\}:1.1), \{hair color\} hair, (\{hair/beard style\}:1.1), neutral expression, wearing dress, (white background:1.4), head horizontally aligned, (uniform lighting:1.4), top of the hair visible, photo for ID. \par} \\[5pt]
Positive Prompt Example & { \raggedright RAW front photo, face portrait photo of (20 years old:1.1), Chilean (female:1.1), dark brown hair, (short haircut:1.1), neutral expression, wearing black suits, (white background:1.4), head horizontally aligned, (uniform lighting:1.4), top of the hair visible, photo for ID. \par} \\ \bottomrule
\end{tabular}
\end{table*}

\begin{figure*}[]
\centering
\includegraphics[scale=0.25]{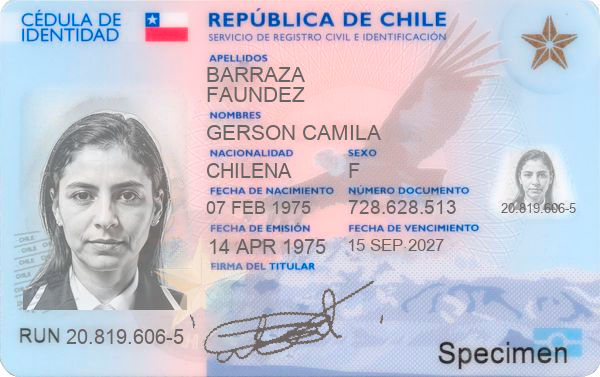}
\includegraphics[scale=0.25]{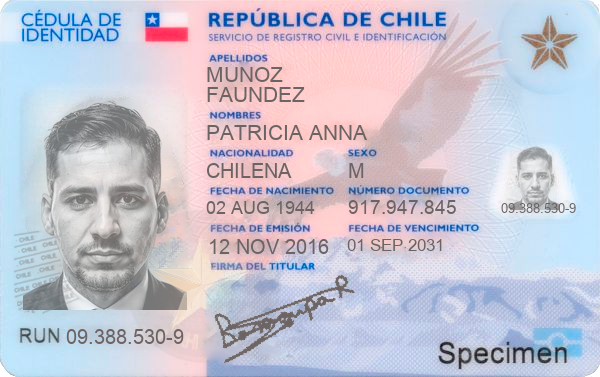}
\includegraphics[scale=0.2]{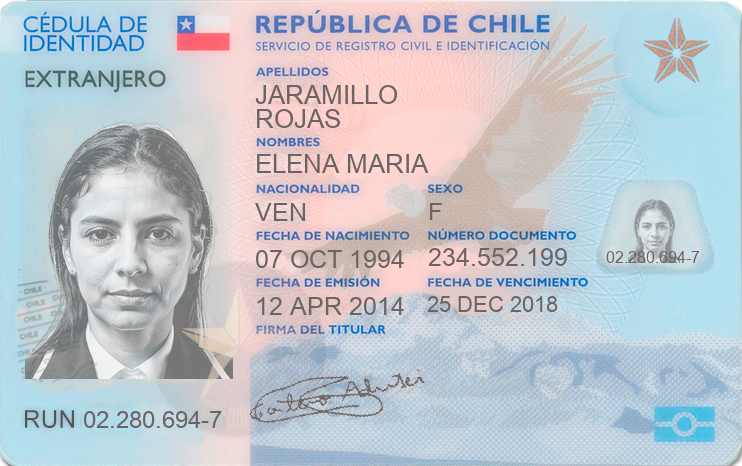}
\caption{Example of a Hybrid ID card generated for citizens and foreign citizens.}
\label{fig:examples-hybrid-IDcards}
\end{figure*}

\subsubsection{Post-processing}
The post-processing procedure primarily involves adjusting the size, transparency, colour space, and orientation of both the generated frontal-facing facial images and the collected signature samples to match the required format based on reference measurements. 

For the frontal-facing facial images, the original inputs are in RGB colour space with a resolution of $512\times512$ pixels. Initially, we utilise Rembg\footnote{\url{https://github.com/danielgatis/rembg}}, an AI tool for rendering the background transparent, to automatically remove the background of the image. Subsequently, each image is converted to grayscale and processed to generate two output variants: one resized to $628 \times 194$ pixels with a transparency level of 65\%, and the other resized to $25 \times 104$ pixels with 50\% transparency. 

For the signature images, the adjustments follow a similar sequence. Given that these images are in grayscale, background removal is achieved by eliminating pixels with values greater than $200$, discarding the white background. Each image is resized to $186\times132$ pixels and rotated $15$ degrees counterclockwise to align with the target layout. 

\subsubsection{Layer composite}
Following the generation and post-processing of individual components—including facial images, textual content, and signatures—the final stage involves composing these elements onto the templates to synthesise complete identity documents. This process contains spatial parameters extracted and stored in the layout JSON files.

The layer composing process is achieved pragmatically, where each component is positioned onto the ID template according to predefined coordinates, size, rotation angles, transparency values, and so on. To implement this composing, we utilise Python with the image processing library PIL and the fake text generation library Faker\footnote{\url{https://faker.readthedocs.io/en/master/}}. PIL allows transformation functions of images, including resizing, rotation, and transparency. Faker renders textual elements using the \enquote{Arial font}, with corresponding text format, font size, and colour that are measured from the reference.

The final version of the generated images based on the hybrid method is illustrated in Figure~\ref{fig:examples-hybrid-IDcards}.

\subsection{One-step-Generation}

Generating an ID card in one-shot is very challenging due to the complexity and density of information present in identity documents, including background patterns, textures with valid information, and the correlation between fields. As a result, one potential solution emerged based on the Large Language Models and fine-tuning the Low-Rank Adaptation~(LoRA)~\cite{hu2022lora}.

LoRA is a method that enables the efficient fine-tuning of large language models (such as CLIP~\cite{cherti2023reproducible} in our case) by introducing a small set of trainable parameters, including $Wq$, $Wk$, $Wv$, and $Wo$, which represent query/key/value/output projection matrices in the self-attention module. Where $W$ or $Wo$ refers to a pretrained weight matrix and $\Delta W$ its accumulated gradient update during adaptation~\cite{hu2022lora}. 

In our proposal, based on Hu et al. approach~\cite{hu2022lora}, instead of updating the entire model weight matrix, the adapter keeps $W$ fixed and learns a low-rank update $\Delta W = AB$. 
While $A$ and $B$ contain trainable parameters. Note that both $Wo$ and $\Delta W$ = $AB$ are multiplied with the same input, and their respective output vectors are summed coordinate-wise.

\begin{figure}[H]
\centering
\includegraphics[scale=0.35]{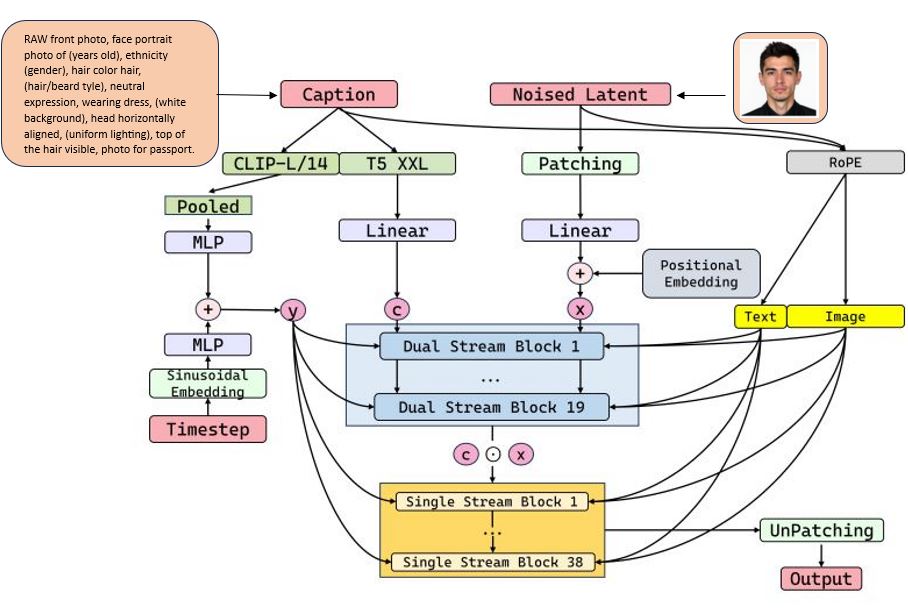}
\caption{Flux architecture based on~\cite{gao2024eraseanything}. The CLIP and T5-XXL blocks are used for fine-tuning.}
\label{fig:flux-archi}
\end{figure}

This approach significantly reduces the number of model parameters.
More specifically, LoRA was used to adjust the attention layers within Dual Stream Blocks, which simultaneously process the textual representations from \enquote{block T5} and Block \enquote{CLIP-L14} alongside the visual representations, as illustrated in Figure~\ref{fig:flux-archi}.

These blocks~(T5 XXL and CLIP-L14) are responsible for enabling effective multimodal interaction between text and image. Additionally, LoRA was also adjusted to train the attention layers in the \enquote{Single Stream Blocks}, which operate after the representations have been merged into a unified signal. 

Due to the need for large amounts of data and the time required to create hybrid document generation through templates, some challenges are still relevant, such as the lack of security codes and elements absence of material; there has been increasing interest in studying and developing models capable of learning and transferring these qualities from bona fide samples into generated simualted bona fide images in one-shot.

\begin{figure}[H]
\centering
\includegraphics[scale=0.13]{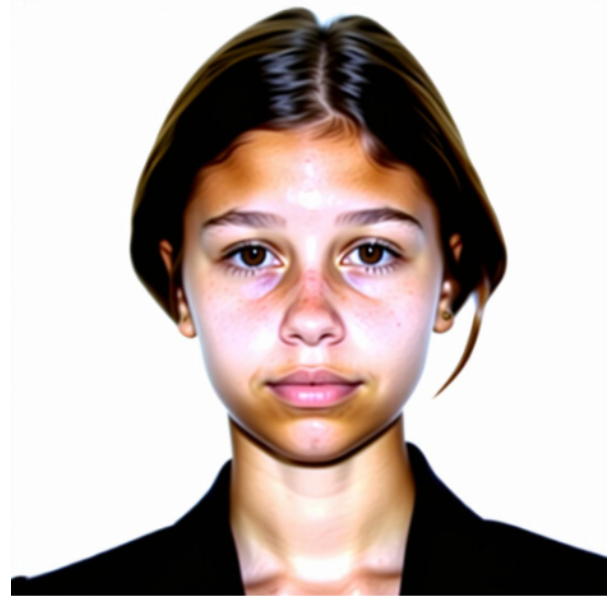}
\includegraphics[scale=0.18]{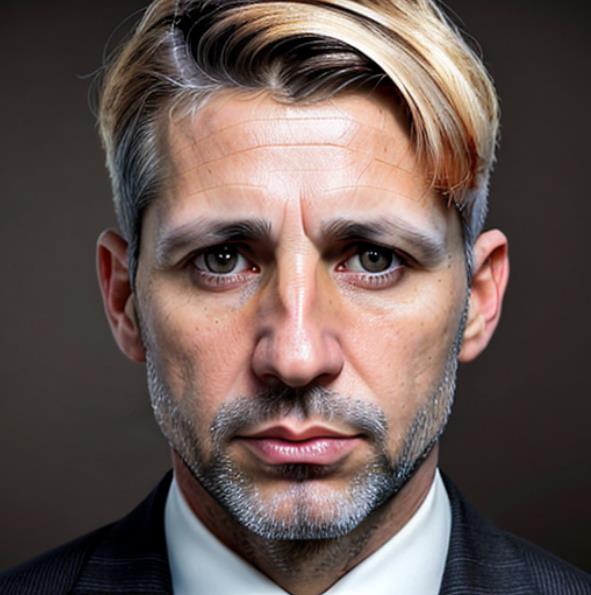}
\includegraphics[scale=0.13]{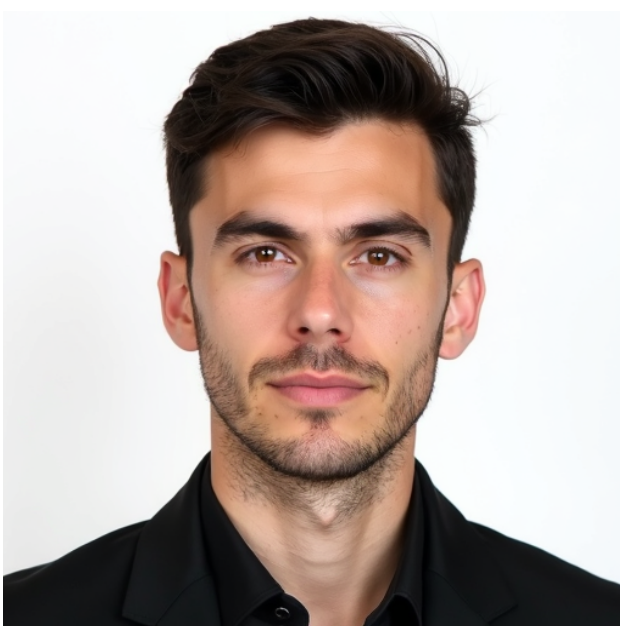}
\caption{Face generation results for each model. Left to right: Stable DiffusionV3, HyperVAE, Flux-1-Dev.}
\label{fig:face-generation}
\end{figure}

In this work, we first evaluated and identified the best face image generation model based on generative models, including Stable-DiffusionV3, Flux-1-dev, and HyperVAE. Afterwards, we extended the use of the best tool to the whole document generation. The best results were obtained by Flux-1-dev, as shown in Figure \ref{fig:face-generation}, which aligns with our previously described requirements in Section~\ref{sec:methods}.

Our approach for one-shot ID card generation is based on Flux-1-Dev combined with LoRA adapter to fine-tune the model and to achieve high consistency in image generation. The main objective is to generate images of Chilean documents. This approach enables more efficient training of large models by utilising fewer steps. Instead of training all the parameters of the original model, LoRA leverages low-dimensional matrices to adjust specific layers of the model.

\subsubsection{Fine-tuning with Gradio-UI}

Flux-1-dev is a model based on a modified U-Net architecture specifically designed for segmentation and feature extraction tasks. This architecture is well known for its encoder-decoder structure, which is composed of convolutional layers and skip connections. To accomplish this task, a tool called AI-Toolkit~\cite{ai_toolkit_ostris} is used, which provides various platforms for performing fine-tuning on this model. 

In our case, we initially used the Gradio-UI interface (Gradio-UI) \cite{abid2019gradio} as a first approach for the fine-tuning process in order to evaluate several prompts and parameters. Gradio-UI is an open-source tool that allows users to create interfaces for testing, visualising, and fine-tuning models without the need for extensive coding. This tool provides a local interface for uploading images intended for fine-tuning, along with a configuration panel to adjust the relevant parameters. 

For the initial attempt, the fine-tuning process was carried out over $10,000$ steps, with a learning rate of $5e-4$, using $150$ \textit{simulated bona fide} images with a fixed resolution of $466\times742$ pixels. However, the results obtained after fine-tuning were lacking coherence and consistency in representing the variable fields and their typography. 

On the other hand, in order to determine the best parameters of the LoRa fine-tuning, a grid search was implemented in the range [2,16]. 

The best parameters selected were: \textit{linear: 8} and \textit{linear alpha: 32}. The linear parameter defines the low-rank used in the matrix decomposition of the layers modified by LoRA. Instead of training the entire weight matrix, LoRA trains two smaller matrices with dimensions \textit{(original dim, 8)} and \textit{(8, original dim)}, which significantly reduces the number of trainable parameters. This approach captures the essential variations without overfitting the LoRA matrix.

Meanwhile, the linear \textit{alpha} parameter acts as a scaling factor. This scaling allows for adjusting the effective influence of LoRA's learning, helping to stabilise training and prevent the lightweight layers from interfering too much.

\begin{figure}[H]
\centering
\includegraphics[scale=0.25]{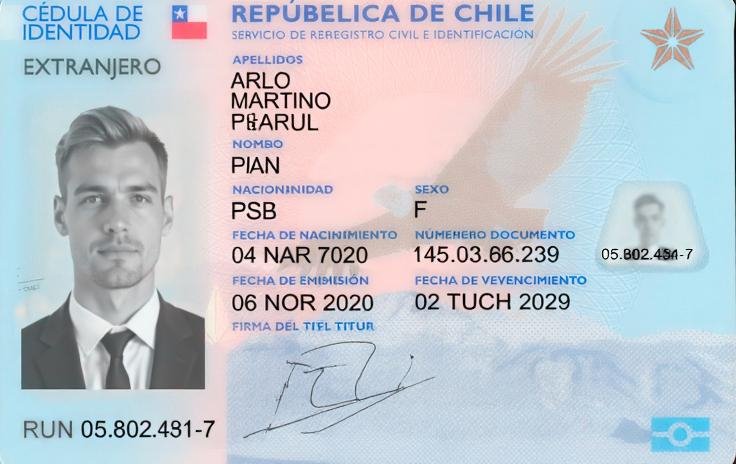}
\caption{Chilean ID document generation results with the Flux-1-dev model fine-tuned with 150 images. \textit{Check the inconsistent names.}} 
\label{fig:chilean_results_flux1}
\end{figure}

For the first attempts, the preliminary results obtained using only 150 images are not satisfactory because the text can not obtain a valid text (demographic information), as is illustrated in Figure \ref{fig:chilean_results_flux1}.  This could be due to the limited number of images used for model adjustment, as the small dataset itself lacks generality and variety.

\subsubsection{Fine-tuning with AI-Toolkit}
Since Gradio-UI imposes certain limitations, such as a maximum number of images for fine-tuning (up to $150$ images), and in order to improve the results, we opted to conduct training without using the interface to avoid these constraints. The model was trained on a local machine using the \textit{AI-Toolkit}~\cite{ai_toolkit_ostris} and a configuration defined in a~\textit{.yml} file, which will be explained in more detail later.

For this second attempt to use LoRA for fine-tuning of the Flux-1-dev model, which generates images in a one-step process, a dataset of $1,100$ images was created, consisting of $1,000$ \textit{simulated bona fide images} using a hybrid method described in Section~\ref{sec:methods}, and $100$ bona fide samples. 

A set of \textit{simulated bona fide} images was supplemented with $100$ bona fide samples, allowing the model to learn about the degradation that document materials undergo over time, as well as the noise present in captures taken in uncontrolled environments with receivers of varying quality. Additionally, for each image, a~\textit{.txt} file with the same name was stored in the same directory, containing the prompt associated with that specific image.

The prompt must be structured so that the model can correctly extract the necessary features for generating the Visual Inspection Zone (VIZ), which is the section of the identity document (i.e. ID card or passport), that contains readable data, which may include static fields and variable fields.) For this reason, the prompt includes several essential parameters for VIZ generation, such as surname, given name, nationality, gender, date of birth, document number, date of issue, and date of expiry.

It is important to emphasise that these variables must be presented in an organised manner within the prompt, each preceded by a static field that clearly indicates the nature of the variable. Only positive prompts were used with Flux-dev-1.

The prompt follows this structure:
\\
\textit{Prompt: Chile ID card with surnames (surnames), names (name), nationality (nationality), gender (gender), date of birth (date of birth), document number (document number), issuance date (issuance date), expiration date (expiration date), RUN (RUN).}
\\
\\
Regarding the configuration parameters, we can highlight a batch size of $1$, a total of $15,000$ steps, a learning rate $(lr)$ of $1e-4$, and input dimensions of $742\times466$ pixels in width and height, respectively.

After completing the fine-tuning, the results obtained were significantly better than those achieved using the Gradio-UI with only $100$ images. The fine-tuning process allowed the model to understand deeper characteristics of the document, such as background textures and the relationship between the prompt and the areas where variable fields should be placed.

Moreover, the model successfully reproduced the background and typography with high similarity and accurately placed the document number over the \textit{CLI field}. The \textit{CLI} is an alphanumeric code that contains the user's information, such as a document number, issuance date, and expiration date. An example that illustrates the results is shown in Figure \ref{fig:chilean_results_flux1}.


\subsection{Visualization}

To assess the similarity results of the \textit{simulated bona fide} samples in comparison to the original bona fide samples, we utilise t-distributed Stochastic Neighbour Embedding (t-SNE)~\cite{t-sne} plots. The t-SNE is a dimensionality reduction technique that maps high-dimensional data into a lower-dimensional space (typically two or three dimensions), allowing for a visual representation of the spatial relationships between data points. This method is particularly useful for identifying clusters or patterns, making it ideal for visually comparing the spatial distribution of original and generated samples.

\begin{figure}[H]
\centering
\includegraphics[scale=0.37]{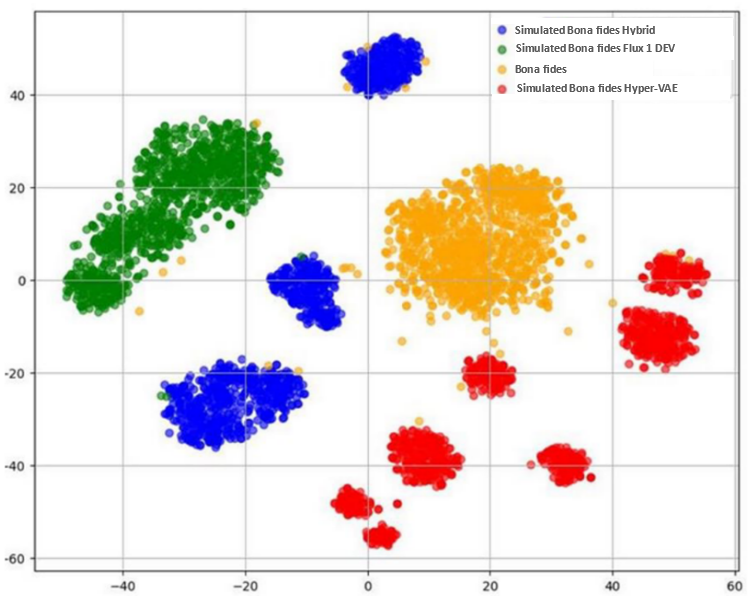}
\caption{t-SNE plot for \textit{simulated bona fide} samples versus bona fide samples.}
\label{fig:tsne-todos}
\end{figure}

Figure~\ref{fig:tsne-todos} illustrates the t-SNE representation of bona fide samples as the yellow cloud and \textit{simulated bona fide} samples using a hybrid approach in blue, a Flux-1-dev approach in green and HyperVAE in the red cloud.

As can be observed, the synthetic samples distribution generated by the Flux-1-dev model (green) is the farthest to the set of bona fide samples, unlike those generated using hybrid and HyperVAE methods. This difference is due to the fact that the dataset used to train Flux-1-dev consisted of 90.91\% ($1,000$) synthetic samples (generated using the proposed hybrid method) and only 9.09\% (100) bona fide images.

\begin{figure}[H]
\centering
\includegraphics[scale=0.42]{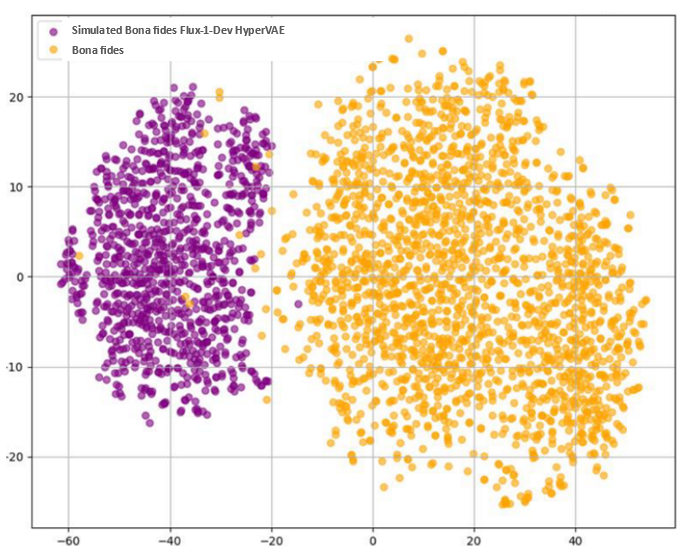}
\caption{t-SNE plot for \textit{simulated bona fide} samples and bona fide samples.}
\label{fig:tsne_todos2}
\end{figure}

To address this limitation, we created a training set consisting entirely of bona fide images (a total of 200) and repeated the fine-tuning process. This new set (Flux-1-dev-2) has successfully improved the previous results and adjusted Flux-1-dev to generate samples that are closer to the real bona fide set. Figure~\ref{fig:tsne_todos2} depicts the t-SNE of the new images generated with Flux-1-dev-2.

After reviewing the t-SNE representation, we proceed to calculate the $FID$ for each set of images generated by the different \textit{simulated bona fide} generation methods in comparison with the real bona fide image set.

\begin{figure*}[]
\centering
\includegraphics[scale=0.45]{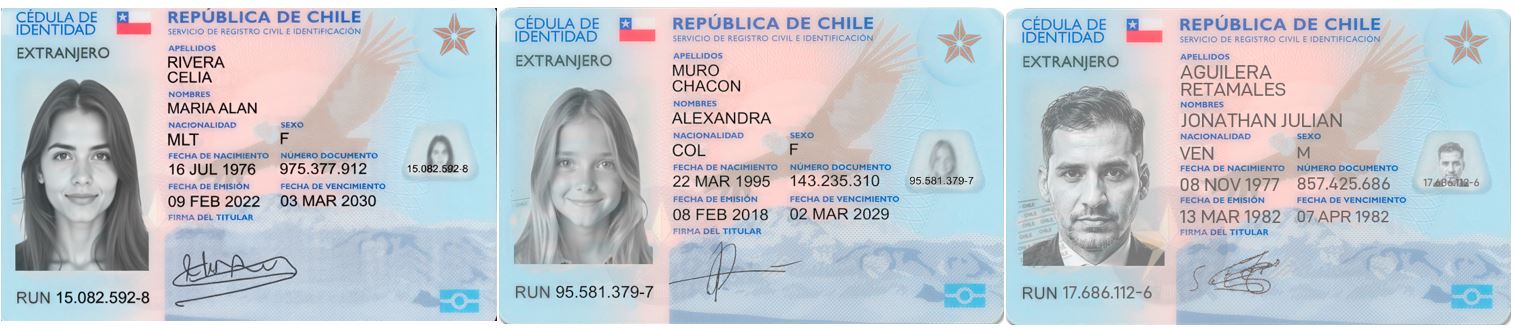}
\caption{Example of three \textit{simulated bona fide} Chilean ID cards images. From left to right. Left: One-step ID card generation by the Flux-1-Dev model fine-tuned with 1,100 images. Middle: Hybrid generation. Right: Hybrid generation using HyperVAE.}
\label{fig:idcards-examples}
\end{figure*}

In Table \ref{tab:FID-scores}, we can observe that the subsets with the lowest $FID$ value are the ones generated in a hybrid manner by HyperVAE, while the set with the highest $FID$ is the one generated in one-step by the Flux-1-dev model. Examples of simulated bona fide ID cards images generated are illustrated in Figure~\ref{fig:idcards-examples}.

\begin{table}[H]
\centering
\caption{FID scores from ID cards for \textit{simulated bona fide}.}
\label{tab:FID-scores}
\begin{tabular}{|c|c|c|c|c|}
\hline
 &
  \begin{tabular}[c]{@{}c@{}} Simulated \\bona fide\\ Flux-1-dev /\\ vs bona fide\end{tabular} &
  \begin{tabular}[c]{@{}c@{}}Simulated\\ bona fide\\ Hybrid /\\ vs bona fide\end{tabular} &
  \begin{tabular}[c]{@{}c@{}}\textbf{Simulated}\\ \textbf{bona fide}\\ \textbf{HyperVAE }/\\  vs bona fide\end{tabular} &
  \begin{tabular}[c]{@{}c@{}}Simulated\\ bona fide\\ Flux-1-dev-2 /\\  vs bona fide\end{tabular}\\ \hline
FID&
  71.75 &
  59.28 &
  \textbf{41.78} &
  53.12 \\ \hline
\end{tabular}%
\end{table}

\vspace{-0.3cm}

\section{Dataset}
\label{sec:data}

As mentioned in the motivation for this work, it is necessary to develop new techniques for improving the generalisation capabilities of the PAD system. 
A new trained dataset based on bona fide samples, screen samples, and printed attack samples was created to evaluate the impact of our \textit{simulated bona fide} images. In addition, different \textit{simulated bona fide} sets of images were generated using the proposed Hybrid and One-step techniques. Table~\ref{tab:summary-sintetic} summarises the synthetic images.

\begin{table}[H]
\centering
\caption{Simulated bona fide generated images and the models used.}
\label{tab:summary-sintetic}
\begin{tabular}{|c|c|c|c|}
\hline
Models & Flux-1-dev (one step) & Hybrid & HyperVAE Hybrid \\ \hline
\begin{tabular}[c]{@{}c@{}}N \\ images\end{tabular} & 1,000       & 1,000   & 1,000      \\ \hline
\end{tabular}%
\end{table}

Consequently, a standard test set was used in order to evaluate the PAD models. This test set consists of $9,329$ images, of which $2,506$ are bona fide images, $1,941$ are screen images, $1,941$ are printed on PVC, and $2,941$ are prints on glossy paper. It is also worth mentioning that this test set includes ID documents from various countries. 
A summary of all subsets used for training the PAD system is shown in Table~\ref{tab:my-table2}. 

\begin{table}[H]
\scriptsize
\centering
\caption{Summary subsets used for training/val/test PAD ID cards system.}
\label{tab:my-table2}
\begin{tabular}{|c|c|c|c|c|}
\hline
 & \begin{tabular}[c]{@{}c@{}}Set\\ Train\end{tabular} & \begin{tabular}[c]{@{}c@{}}Set\\ Val\end{tabular} & \begin{tabular}[c]{@{}c@{}}Set\\ Test\end{tabular} & \begin{tabular}[c]{@{}c@{}}\textbf{Total}\end{tabular} \\ \hline
Bona fide & 9,362 & 2,348 & 2,506 & 14,216\\ \hline
Screen   & 9,362 & 2,348 & 1,941 & 13,651\\ \hline
PVC      & 4,607 & 1,157 & 1,941 & 7,705\\ \hline
Print    & 9,362 & 2,348 & 2,941 & 14,651\\ \hline
\textbf{Total}    & 32,693 & 8,201 & 9,329 & 50,223\\ \hline
\end{tabular}
\end{table}

\section{Experiments and Results}
\label{sec:results}

\subsection{PAD classification}

A series of experiments was conducted to evaluate the effect of  \textit{simulated bona fide} samples in the training process and the positive impact on various methods on a PAD ID card system. 

The evaluations were divided into three experiments. These experiments are based on a PAD ID card system baseline that uses an EfficientNetV2 architecture with a B3-base model. This specific network was chosen because it has yielded one of the best results in the PAD state-of-the-art~\cite{Tapia-IJCB2024}. Moreover, throughout this research, one of the focuses has been on evaluating the trade-off between the quality of the \textit{simulated bona fide} approach generated and the network's performance. 

The parameters used to train the network are: an input size of $384\times384$, pretrained weights from ImageNet22K, a learning rate of $5e-5$, $150$ epochs, $200$ steps per epoch, and an Adam optimiser. These parameters are consistent across all experiments, as the goal is to study the substitution or complement of bona fide samples.

\subsection{Experiment 1 - Baseline}
A PAD ID card system was trained as a baseline to compare and evaluate the images generated by both synthetic methods.

For this experiment, we selected $9,362$ bona fide images and $23,331$ attack images, separated into printed and screen attack images for training. This baseline consists of an EfficientNetV2-B3 \cite{efficientnet} network. The results of the baseline, tested on real data, are presented in Table \ref{tab:my-table3}, which shows an EER of 9.27\%.

\begin{table}[H]
\centering
\caption{Summary PAD Results for the Baseline – Experiment 1.}
\label{tab:my-table3}
\begin{tabular}{|c|c|}
\hline
Models   & Baseline \\ \hline
\textbf{EER(\%) }         & \textbf{9.27}              \\ \hline
BPCER10(\%)      & 8.46              \\ \hline
BPCER20(\%)      & 18.05             \\ \hline
BPCER100(\%)     & 40.87             \\ \hline
\end{tabular}
\begin{minipage}{0.9\linewidth}
\centering
\end{minipage}
\end{table}
\vspace{-0.5cm}

\subsection{Experiment 2 - adding simulated bona } 

For this experiment, we complement the bona fide samples with each \textit{simulated bona fide} method. In this case, the $14,216$ bona fide images from the baseline model will be augmented with $1,000$ images from \textit{simulated bona fide} by Flux-1-dev. 
In the next round, $1,000$ samples were generated using hybrid generation. Afterward, in the next round, $1,000$ samples were generated with HyperVAE. Finally, we generated $1,000$ samples with Flux-1-dev-2 (a model fine-tuned using only genuine samples). 

It should be noted that the number of print and screen attack images has not been modified, so there are still $36,000$ images across the types of attacks. The summary results are shown in Table \ref{tab:my-table1}. The best results are highlighted in bold black.

\begin{table}[H]
\centering
\caption{Baseline dataset + simulated bona fide samples using a hybrid approach with Flux-1-dev and with HyperVAE, respectively. All the results are in (\%).}
\label{tab:my-table1}
\begin{tabular}{|c|c|c|c|c|}
\hline
Models &
  \begin{tabular}[c]{@{}c@{}}Baseline + \\ simulated \\ bona fide\\ Flux-1 \\ dev) \\ (one-step)\end{tabular} &
  \begin{tabular}[c]{@{}c@{}}Baseline + \\ simulated \\ bona fide \\ (Hybrid)\end{tabular} &
  \begin{tabular}[c]{@{}c@{}}Baseline + \\ simulated \\ \textbf{bona fide}\\ (\textbf{HyperVAE} \\ \textbf{Hybrid})\end{tabular} & 
  \begin{tabular}[c]{@{}c@{}}Baseline + \\ simulated\\ bona fide\\ (Flux-1 \\dev-2) \\ (one-step)\end{tabular}
  \\ \hline
EER      & 10.98 & 14.36 & \textbf{8.26} & 10.82\\ \hline
BPCER10  & 12.19 & 19.29 & \textbf{5.90}  & 11.83\\ \hline
BPCER20  & 24.34 & 28.76 & \textbf{14.88} & 22.62\\ \hline
BPCER100 & 45.48 & 49.09 & \textbf{40.06} & 46.28\\ \hline
\end{tabular}%
\end{table}

\subsection{Experiment 3 - simulated bona fide replacement}

For this experiment, in order to analyse the impact of our approach, we replaced the genuine bona fide samples with \textit{simulated bona fide}. In this case, the $1,000$ bona fide images from the baseline will be replaced by $1,000$ images \textit{simulated bona fide} by Flux-1-dev, $1,000$ samples from the hybrid generation, and finally, $1,000$ samples generated by HyperVAE, respectively. It should be noted that the number of printed and screened attack images has not been altered, so there are still $36,000$ images between print and screen attacks. 

Table \ref{tab:my-table4} shows the summary results for experiment 3. The lowest EER was obtained for the one-step model, which is the best-performing model in this experiment. However, with a modest detection accuracy. As expected, a large number of images is necessary to train a PAD and obtain generalisation capabilities.
\vspace{-0.3cm}

\begin{table}[H]
\scriptsize
\centering
\caption{Summary results with the substitution of 1,000 bona fide samples with \textit{simulated bona fide}. All the results are in (\%).}
\label{tab:my-table4}
\begin{tabular}{|c|c|c|c|c|c|}
\hline
Models &
  \begin{tabular}[c]{@{}c@{}}Bona fide \\ (1000 \\ samples)\\ \end{tabular} &
  \begin{tabular}[c]{@{}c@{}}Simulated \\ Bona fide\\ Flux-1 \\dev \\ (one step)\end{tabular} &
  \begin{tabular}[c]{@{}c@{}}Simulated\\ bona fide\\ (Hybrid)\end{tabular} &
  \begin{tabular}[c]{@{}c@{}}Simulated\\ Bona fide\\ HyperVAE \\ (Hybrid)\end{tabular} &
  \begin{tabular}[c]{@{}c@{}}Simulated \\ Bona fide\\ Flux-1 \\dev-2 \\ (one step)\end{tabular}
    \\ \hline
EER  & 17.76 & \textbf{33.66} & 35.31 & 47.45 & 39.48\\ \hline
BPCER10 & 29.56 & \textbf{69.23} & 71.40 & 85.51 & 74.80\\ \hline
BPCER20 & 44.16 & \textbf{80.54} & 81.78 & 91.13 & 84.87\\ \hline
BPCER100& 67.66 & \textbf{93.06} & 95.06 & 97.07 & 95.94\\ \hline
\end{tabular}
\end{table}

\subsection{Experiment 4 - Only 1000 samples per class} 

In Experiment 4, to evaluate the influence of \textit{simulated bona fide} samples considering an equilibrated dataset, the number of print and screen samples fed into EfficientNet-V2 has also been reduced to $1,000$. This reduction enables us to assess the impact of synthetic samples on the model's performance more clearly. Additionally, due to the smaller dataset size, the number of steps and epochs has also been decreased. Specifically, the training was conducted over $100$ epochs with $94$ steps per epoch and a learning rate of $5e-4$.

Table \ref{tab:my-table_exp_4} shows a summary of the results with the new \textit{simulated bona fide} images generated in a PAD system trained from scratch. The results obtained are better than Experiment 3, but only with a reduced set of images. The HyperVAE hybrid method obtained the lowest EER.

\begin{table}[H]
\scriptsize
\centering
\caption{Summary results between bona fide versus \textit{simulated bona fide} ID cards. All the results are in (\%).}
\label{tab:my-table_exp_4}
\begin{tabular}{|c|c|c|c|c|c|}
\hline
Models &
  \begin{tabular}[c]{@{}c@{}}Bona fide \\ (1000 \\ samples)\end{tabular} &
  \begin{tabular}[c]{@{}c@{}}Simulated \\ bona fide\\ Flux-1-dev\\ (one step)\end{tabular} &
  \begin{tabular}[c]{@{}c@{}}Simulated\\ bona fide\\ (Hybrid)\end{tabular} &
  \begin{tabular}[c]{@{}c@{}}Simulated\\ bona fide\\ HyperVAE\\ (Hybrid)\end{tabular} &
  \begin{tabular}[c]{@{}c@{}}Simulated\\ bona fide\\ Flux-1-dev \\ 2\\ (one step)\end{tabular} \\ \hline
ERR    & 11.35 & 41.43 & 27.05 & \textbf{24.30} & 27.45\\ \hline
BPCER10 & 13.90 & 81.60 & 62.78 & \textbf{44.66} & 50.70\\ \hline
BPCER20 & 27.95 & 88.62 & 77.11 & \textbf{67.42} & 68.68\\ \hline
BPCER100 & 65.03 & 97.19 & 91.99 & \textbf{91.99} & 89.89\\ \hline
\end{tabular}%
\end{table}

\subsection{COTS evaluation}
To evaluate the impact of our \textit{simulated bona fide} Chilean ID card images in a commercial PAD system and to establish their reliability, we performed a third-party assessment. 
A commercial-off-the-shelf (COTS) algorithm was utilised in collaboration with a private company, which provided us with a classification score for each image per method. 

This system was evaluated using three main sets of \textit{simulated bona fide} generated documents as an evaluation set. Each consisting of 1,000 images, utilising Flux-1-dev, a hybrid approach, and the hybrid method HyperVAE, separately. Additionally, 1,000 print and 1,000 screen ID card images were used.

\begin{figure}[H]
\centering
\includegraphics[scale=0.45]{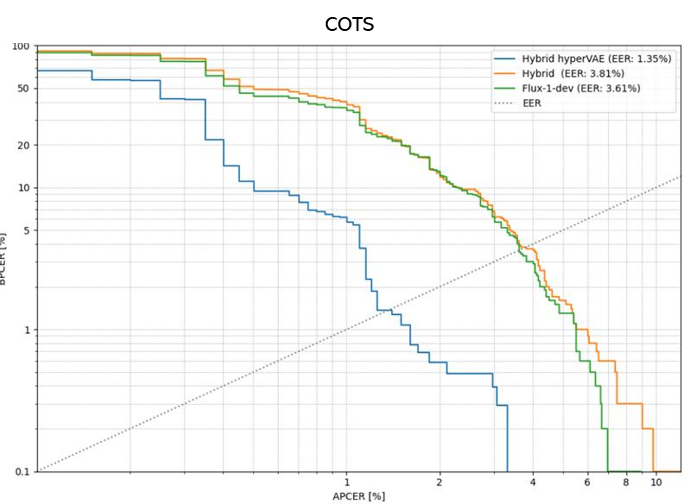}
\caption{DET curve generated by Flux-1-dev VS Hybrid VS Hybrid HyperVAE.}
\label{fig:DET}
\end{figure}

Figure \ref{fig:DET} illustrates the DET curves, demonstrating the performance of the COTS-PAD against the out-of-set for each method, respectively. 
Since all sets include the same print and screen attacks, the performance differences can be directly attributed to the nature of the \textit{simulated bona fide} images.

The curve corresponding to the hybrid samples generated using the hybrid method achieved lower performance from our perspective, as it obtained an EER of 3.81\%, indicating that this image set poses a greater challenge for the system, as \textit{simulated bona fide} images were detected as bona fide (Our goal). 

The curve related to the images generated with Flux-1-dev shows a slight improvement, with an EER of 3.61\%, suggesting that these images are slightly less detectable.  The best results (Lower EER) were obtained by HyperVAE methods, indicating that the PAD system is less effective against this type of \textit{simulated bona fide}. 

In all the curves, the position further away from the origin, along with the BPCER and APCER values, reflects the error rate and the ability of the system to discriminate between our \textit{simulated bona fide} samples and attacks.

It is essential to highlight that COTS systems do not contain synthetic images in the trained set as bona fide.

\section{Conclusion}
\label{sec:conclusion}

After extensive experimentation, several key conclusions can be made. Firstly, the generation of \textit{simulated bona fide} ID card images opens a new insight into image generation because we can also train the system to mimic the distribution of bona fide ID cards. Moreover, a significant opportunity to generate data in a restricted domain, but also a challenge, because today there are several open-access platforms that make it \textit{easy} to create images that can fool the PAD system.

In face generation with a focus on ID cards, Flux-1-dev outperforms Stable-DiffusionV3 and Realistic-HyperVAE, effectively meeting requirements with detailed prompts and without requiring LoRA model fine-tuning.

For document generation, fine-tuning Flux-1-dev's LoRA model with a well-structured prompt and high-quality images enhances its ability to accurately represent both document content and background. However, the best results are achieved with the Hybrid model, which combines the HyperVAE model with layer-based compositing methods.

Additionally, \textit{simulated bona fide} document samples can enhance training datasets, improving the performance of the EfficientNet model and potentially replacing real samples with minimal error. However, the model has been trained on a limited number of synthetic samples, indicating that a larger and more diverse set of bona fide images may be necessary to yield better results. This should be our future work. 

Overall, all these techniques are fully reproducible based on JSON files and GitHub repositories (Upon acceptance).


\section*{Acknowledgements}
This work was partially supported by Facephi project PIDI and the EU Horizon under G.A. EINSTEIN (101121280) and CarMen (101168325), and the German Federal Ministry of Education and Research and the Hessian Ministry of Higher Education, Research, Science and the Arts within their joint support of the National Research Center for Applied Cybersecurity ATHENE.

\bibliographystyle{IEEEtran}
\bibliography{sample.bib}

\begin{thebibliography}{10}
\providecommand{\url}[1]{#1}
\csname url@samestyle\endcsname
\providecommand{\newblock}{\relax}
\providecommand{\bibinfo}[2]{#2}
\providecommand{\BIBentrySTDinterwordspacing}{\spaceskip=0pt\relax}
\providecommand{\BIBentryALTinterwordstretchfactor}{4}
\providecommand{\BIBentryALTinterwordspacing}{\spaceskip=\fontdimen2\font plus
\BIBentryALTinterwordstretchfactor\fontdimen3\font minus \fontdimen4\font\relax}
\providecommand{\BIBforeignlanguage}[2]{{%
\expandafter\ifx\csname l@#1\endcsname\relax
\typeout{** WARNING: IEEEtran.bst: No hyphenation pattern has been}%
\typeout{** loaded for the language `#1'. Using the pattern for}%
\typeout{** the default language instead.}%
\else
\language=\csname l@#1\endcsname
\fi
#2}}
\providecommand{\BIBdecl}{\relax}
\BIBdecl

\bibitem{Karras2019stylegan2}
T.~Karras, S.~Laine, M.~Aittala, J.~Hellsten, J.~Lehtinen, and T.~Aila, ``Analyzing and improving the image quality of {StyleGAN},'' in \emph{Proc. CVPR}, 2020.

\bibitem{stylegan3}
T.~Karras, M.~Aittala, S.~Laine, E.~H\"ark\"onen, J.~Hellsten, J.~Lehtinen, and T.~Aila, ``Alias-free generative adversarial networks,'' in \emph{Proc. NeurIPS}, 2021.

\bibitem{MIDV500}
V.~Arlazarov, K.~Bulatov, T.~Chernov, and V.~Arlazarov, ``{MIDV-500}: a dataset for identity document analysis and recognition on mobile devices in video stream,'' \emph{Computer Optics}, vol.~43, pp. 818--824, 10 2019.

\bibitem{MIDV2020}
B.~Konstantin, E.~E, T.~Daniil, S.~Natalya, C.~Yulia, S.~Alexander, U.~S, M.~Zuheng, B.~Jean-Christophe, L.~Muzzamil, and A.~Vladimir, ``{MIDV-2020}: a comprehensive benchmark dataset for identity document analysis,'' \emph{Computer Optics}, vol.~46, pp. 252--270, 03 2022.

\bibitem{MIDVDLC2021}
E.~D. Polevoy Dimitry~and, Sigareva Irina~and, A.~Vladimir, N.~Dmitry, M.~Zuheng, L.~Muhammad, and B.~Jean-Christophe, ``{D}ocument {L}iveness {C}hallenge {DLC-2021} - part 1 (or, cg),'' May 2022.

\bibitem{KID34K}
E.-J. Park, S.-Y. Back, J.~Kim, and S.~S. Woo, ``{KID34K}: A dataset for online identity card fraud detection,'' in \emph{Proceedings of the 32nd ACM International Conf. on Information and Knowledge Management}, ser. CIKM '23.\hskip 1em plus 0.5em minus 0.4em\relax New York, NY, USA: Association for Computing Machinery, 2023, p. 5381–5385.

\bibitem{IDNet}
L.~Xie, Y.~Wang, H.~Guan, S.~Nag, R.~Goel, N.~Swamy, Y.~Yang, C.~Xiao, J.~Prisby, R.~MacIejewski, and J.~Zou, ``{IDNet}: A novel identity document dataset via few-shot and quality-driven synthetic data generation,'' in \emph{Proceedings - IEEE Intl. Conf. on Big Data, BigData}.\hskip 1em plus 0.5em minus 0.4em\relax Institute of Electrical and Electronics Engineers Inc., 2024, pp. 2244--2253.

\bibitem{GONZALEZ-PR}
S.~Gonzalez and J.~E. Tapia, ``Forged presentation attack detection for {ID} cards on remote verification systems,'' \emph{Pattern Recognition}, vol. 162, p. 111352, 2025.

\bibitem{benalcazar}
D.~Benalcazar, J.~Tapia, S.~Gonzalez, and C.~Busch, ``Synthetic {ID} card image generation for improving presentation attack detection,'' \emph{IEEE Transactions on Information Forensics and Security}, vol.~18, pp. 1814--1824, 2023.

\bibitem{Markham}
R.~P. Markham, J.~M.~E. López, M.~Nieto-Hidalgo, and J.~E. Tapia, ``{Open-Set}: {ID} card presentation attack detection using neural style transfer,'' \emph{IEEE Access}, vol.~12, pp. 68\,573--68\,585, 2024.

\bibitem{isola2017image}
P.~Isola, J.-Y. Zhu, T.~Zhou, and A.~A. Efros, ``Image-to-image translation with conditional adversarial networks,'' in \emph{IEEE Conf. on Computer Vision and Pattern Recognition (CVPR)}, 2017.

\bibitem{Tapia-IJCB2024}
J.~E. Tapia, N.~Damer, C.~Busch, J.~M. Espin, J.~Barrachina, A.~S. Rocamora, K.~Ocvirk, L.~Alessio, B.~Batagelj, S.~Patwardhan, R.~Ramachandra, R.~Mudgalgundurao, K.~Raja, D.~Schulz, and C.~Aravena, ``First competition on presentation attack detection on id card,'' in \emph{IEEE Intl. Joint Conf. on Biometrics (IJCB)}, 2024, pp. 1--10.

\bibitem{Chen-tifs}
C.~Chen, S.~Zhang, F.~Lan, and J.~Huang, ``Domain-agnostic document authentication against practical recapturing attacks,'' \emph{IEEE Transactions on Information Forensics and Security}, vol.~17, pp. 2890--2905, 2022.

\bibitem{SD3}
P.~Esser, S.~Kulal, A.~Blattmann, R.~Entezari, J.~M\"{u}ller, H.~Saini, Y.~Levi, D.~Lorenz, A.~Sauer, F.~Boesel, D.~Podell, T.~Dockhorn, Z.~English, and R.~Rombach, ``Scaling rectified flow transformers for high-resolution image synthesis,'' in \emph{41st Intl. Conf. on Machine Learning}, ser. ICML'24.\hskip 1em plus 0.5em minus 0.4em\relax JMLR.org, 2024.

\bibitem{bitflux}
\BIBentryALTinterwordspacing
C.~Yang, C.~Liu, X.~Deng, D.~Kim, X.~Mei, X.~Shen, and L.-C. Chen, ``1.58-bit flux,'' 2024. [Online]. Available: \url{https://arxiv.org/abs/2412.18653}
\BIBentrySTDinterwordspacing

\bibitem{gao2024eraseanything}
D.~Gao, S.~Lu, S.~Walters, W.~Zhou, J.~Chu, J.~Zhang, B.~Zhang, M.~Jia, J.~Zhao, Z.~Fan \emph{et~al.}, ``Eraseanything: Enabling concept erasure in rectified flow transformers,'' \emph{ICML 2025}, 2024.

\bibitem{nguyen}
P.~Nguyen, T.~Tran, S.~Gupta, S.~Rana, H.-C. Dam, and S.~Venkatesh, ``Variational {Hyper-encoding} networks,'' in \emph{Machine Learning and Knowledge Discovery in Databases. Research Track}.\hskip 1em plus 0.5em minus 0.4em\relax Springer Intl. Publishing, 2021, pp. 100--115.

\bibitem{MIDV2019}
K.~Bulatov, D.~Matalov, and V.~Arlazarov, ``{MIDV-2019}: Challenges of the modern mobile-based document ocr,'' 10 2019.

\bibitem{MIDVHolo}
L.~Koliaskina, E.~Emelianova, D.~Tropin, V.~Popov, K.~Bulatov, D.~Nikolaev, and V.~Arlazarov, ``{MIDV-Holo}: A dataset for id document hologram detection in a video stream,'' in \emph{Intl. Conf. on Document Analysis and Recognition}.\hskip 1em plus 0.5em minus 0.4em\relax Springer, 2023, pp. 486--503.

\bibitem{SIDTD}
C.~Boned, M.~Talarmain, N.~Ghanmi, G.~Chiron, S.~Biswas, A.~M. Awal, and O.~R. Terrades, ``Synthetic dataset of {ID} and travel document,'' 2024.

\bibitem{FMIDV}
M.~Al-Ghadi, Z.~Ming, P.~Gomez-Krämer, J.-C. Burie, M.~Coustaty, and N.~Sidere, ``Guilloche detection for {ID} authentication: A dataset and baselines,'' in \emph{IEEE 25th Intl. Workshop on Multimedia Signal Processing (MMSP)}, 2023, pp. 1--6.

\bibitem{Gonzalez-Tbiom}
S.~Gonzalez, A.~Valenzuela, and J.~Tapia, ``Hybrid two-stage architecture for tampering detection of chipless {ID} cards,'' \emph{IEEE Transactions on Biometrics, Behavior, and Identity Science}, vol.~3, no.~1, pp. 89--100, 2021.

\bibitem{fid}
Y.~Yu, W.~Zhang, and Y.~Deng, ``Frechet inception distance {(FID)} for evaluating gans,'' 09 2021.

\bibitem{vargas2007off}
F.~Vargas, M.~Ferrer, C.~Travieso, and J.~Alonso, ``Off-line handwritten signature {GPDS-960} corpus,'' in \emph{9th Intl. Conf. on Document Analysis and Recognition (ICDAR 2007)}, vol.~2.\hskip 1em plus 0.5em minus 0.4em\relax IEEE, 2007, pp. 764--768.

\bibitem{onot}
N.~Di~Domenico, G.~Borghi, A.~Franco, D.~Maltoni \emph{et~al.}, ``{ONOT}: a high-quality {ICAO}-compliant synthetic mugshot dataset,'' in \emph{The 18th IEEE Intl. Conf. on Automatic Face and Gesture Recognition (FG)}, 2024, pp. 1--6.

\bibitem{rombach2022high}
R.~Rombach, A.~Blattmann, D.~Lorenz, P.~Esser, and B.~Ommer, ``High-resolution image synthesis with latent diffusion models,'' in \emph{Proceedings of the IEEE/CVF Conf. on Computer Vision and Pattern Recognition}, 2022, pp. 10\,684--10\,695.

\bibitem{Merkle-OFIQ-Report-240930}
J.~Merkle, C.~Rathgeb, B.~Herdeanu, B.~Tams, D.~Lou, A.~D\"orsch, M.~Schaubert, J.~Dehen, L.~Chen, X.~Yin, D.~Huang, A.~Stratmann, M.~Ginzler, M.~Grimmer, and C.~Busch, ``Open source face image quality ({OFIQ}) - implementation and evaluation of algorithms,'' \url{https://www.bsi.bund.de/SharedDocs/Downloads/EN/BSI/OFIQ/Projektabschlussbericht_OFIQ_1_0.pdf}, September 2024, last accessed: 2025-02-22.

\bibitem{hu2022lora}
\BIBentryALTinterwordspacing
E.~J. Hu, yelong shen, P.~Wallis, Z.~Allen-Zhu, Y.~Li, S.~Wang, L.~Wang, and W.~Chen, ``Lo{RA}: Low-rank adaptation of large language models,'' in \emph{Intl. Conf. on Learning Representations}, 2022. [Online]. Available: \url{https://openreview.net/forum?id=nZeVKeeFYf9}
\BIBentrySTDinterwordspacing

\bibitem{cherti2023reproducible}
M.~Cherti, R.~Beaumont, R.~Wightman, M.~Wortsman, G.~Ilharco, C.~Gordon, C.~Schuhmann, L.~Schmidt, and J.~Jitsev, ``Reproducible scaling laws for contrastive language-image learning,'' in \emph{IEEE/CVF Conf. on Computer Vision and Pattern Recognition}, 2023, pp. 2818--2829.

\bibitem{ai_toolkit_ostris}
J.~Burkett, ``{AI Toolkit}: The ultimate training suite for finetuning diffusion models,'' \url{https://github.com/ostris/ai-toolkit}, 2025, versión 1.0, licencia MIT.

\bibitem{abid2019gradio}
A.~Abid, A.~Abdalla, A.~Abid, D.~Khan, A.~Alfozan, and J.~Zou, ``Gradio: Hassle-free sharing and testing of ml models in the wild,'' \url{https://github.com/gradio-app/gradio}, 2019, versión 5.x, licencia Apache 2.0.

\bibitem{t-sne}
L.~van~der Maaten and G.~Hinton, ``Visualizing data using t-sne,'' \emph{Journal of Machine Learning Research}, vol.~9, pp. 2579--2605, 2008.

\bibitem{efficientnet}
M.~Tan and Q.~Le, ``{E}fficient{N}et: Rethinking model scaling for convolutional neural networks,'' in \emph{36th Intl. Conf. on Machine Learning}, ser. Proceedings of Machine Learning Research, K.~Chaudhuri and R.~Salakhutdinov, Eds., vol.~97.\hskip 1em plus 0.5em minus 0.4em\relax PMLR, 2019, pp. 6105--6114.

\end{thebibliography}
\vspace{-0.3cm}
%
\begin{IEEEbiography}[{\includegraphics[width=1in,height=1.25in,clip,keepaspectratio]{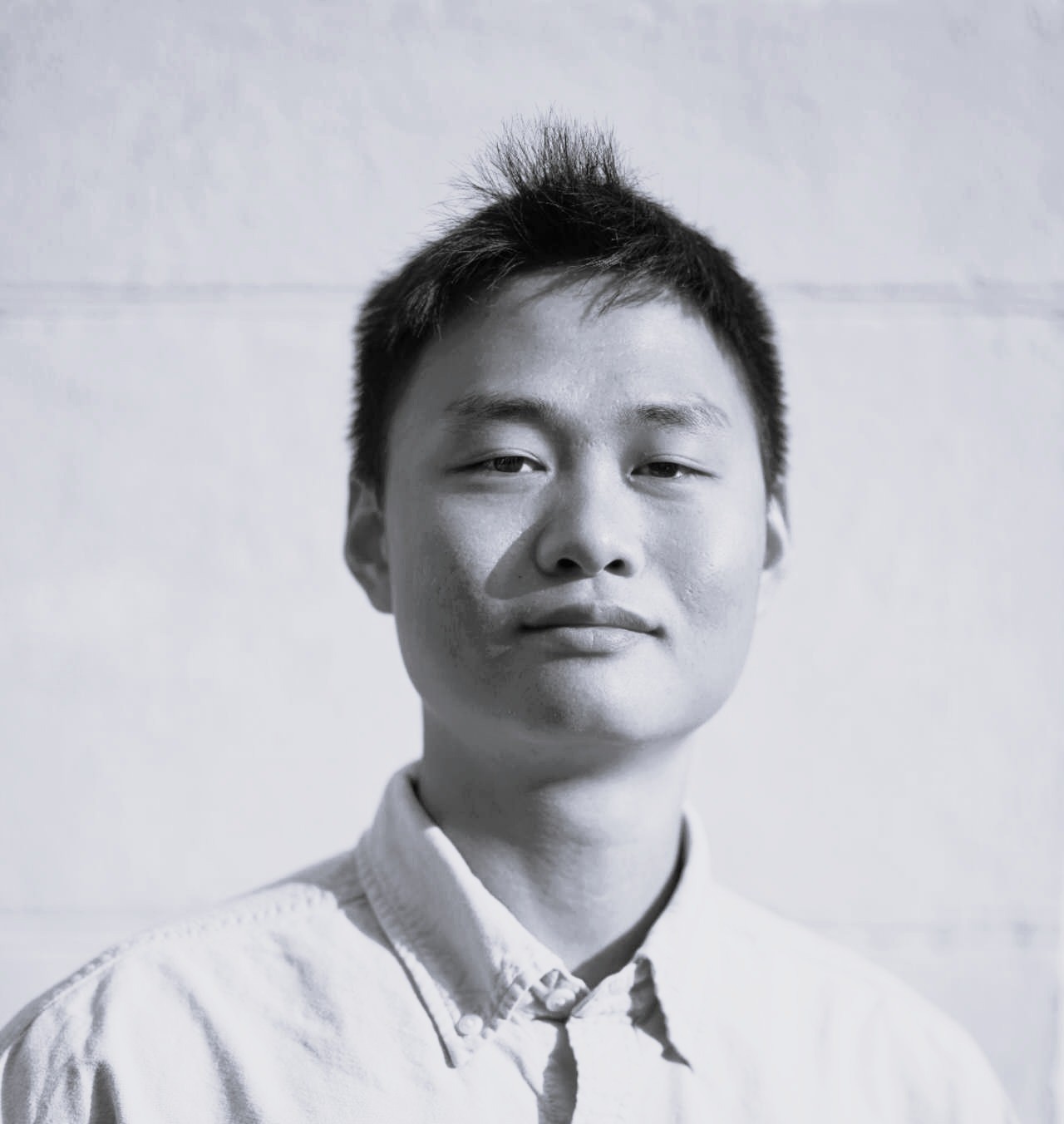}}]{Qingweng Zeng} received a B.Sc. degree in Data Science from the Institute of Disaster Prevention, China, in 2023. He is pursuing his M.Sc. degree in Human-Centred Artificial Intelligence at the Technical University of Denmark.  His research interests include the application of generative models in the field of biometrics.
\end{IEEEbiography}
\vspace{-0.5cm}

\begin{IEEEbiography}[{\includegraphics[width=0.9in,height=1.25in,clip,keepaspectratio]{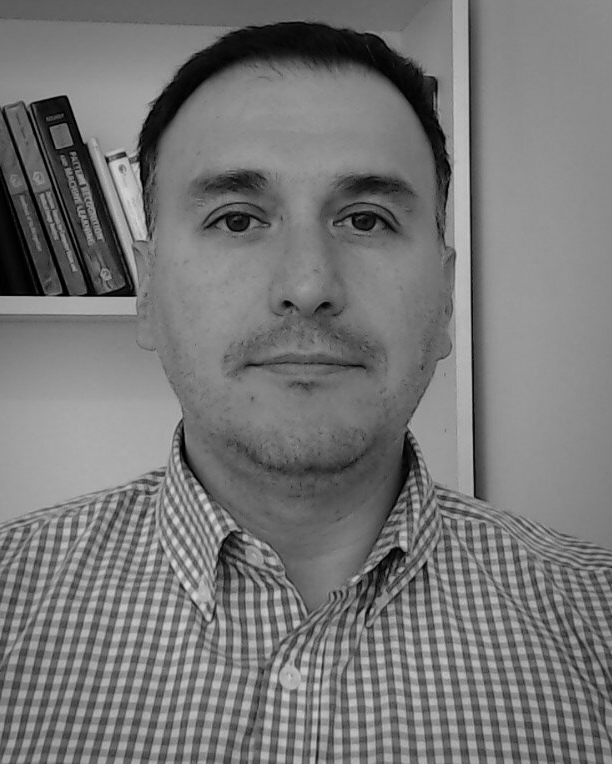}}]{Juan Tapia} received the P.E. degree in electronics engineering from Universidad Mayor, in 2004, and the M.S. and Ph.D. degrees in electrical engineering from the Department of Electrical Engineering, Universidad de Chile, in 2012 and 2016, respectively. In addition, he spent one year of internship with the University of Notre Dame. In 2016, he received the Award for Best Ph.D. Thesis. From 2016 to 2017, he was an Assistant Professor at Universidad Andrés Bello. From 2018 to 2020, he was the Research and Development Director for the electricity and electronics area with INACAP, Universidad Tecnológica de Chile, the Research and Development Director of TOC Biometrics Company, and an International Advisor on biometrics for face, iris applications and forensic/tampering ID-card detection. He is currently an Entrepreneur and a Senior Researcher with Hochschule Darmstadt (H-DA), leading EU projects, such as iMARS, EINSTEIN and CarMen. His main research interests include pattern recognition and deep learning applied to iris biometrics, morphing, feature fusion, and feature selection. He serves as a reviewer for several journals and conferences. He is on behalf of the German DIN as a Member of the ISO/IEC Sub-Committee 37 on biometrics.
\end{IEEEbiography}
\vspace{-0.5cm}

\begin{IEEEbiography}[{\includegraphics[width=1in,height=1.25in, clip,keepaspectratio]{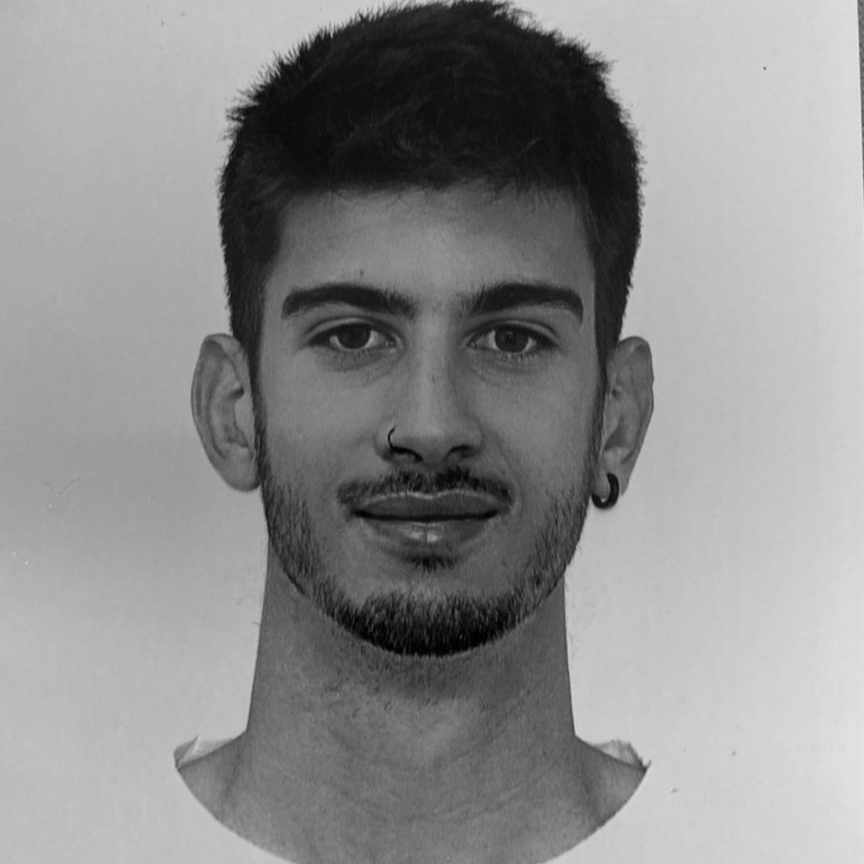}}]{Izan García} received a B.Sc. degree in Computer Engineering from the University of Alicante in 2022, and a Master's degree in Data Science from the University of Alicante in 2024. Currently, he is working as a junior researcher at Facephi company. The main interest topics are related to synthetic image generation and Presentation attack systems.
\end{IEEEbiography}
\vspace{-0.5cm}

\begin{IEEEbiography}[{\includegraphics[width=1in,height=1.25in,clip,keepaspectratio]{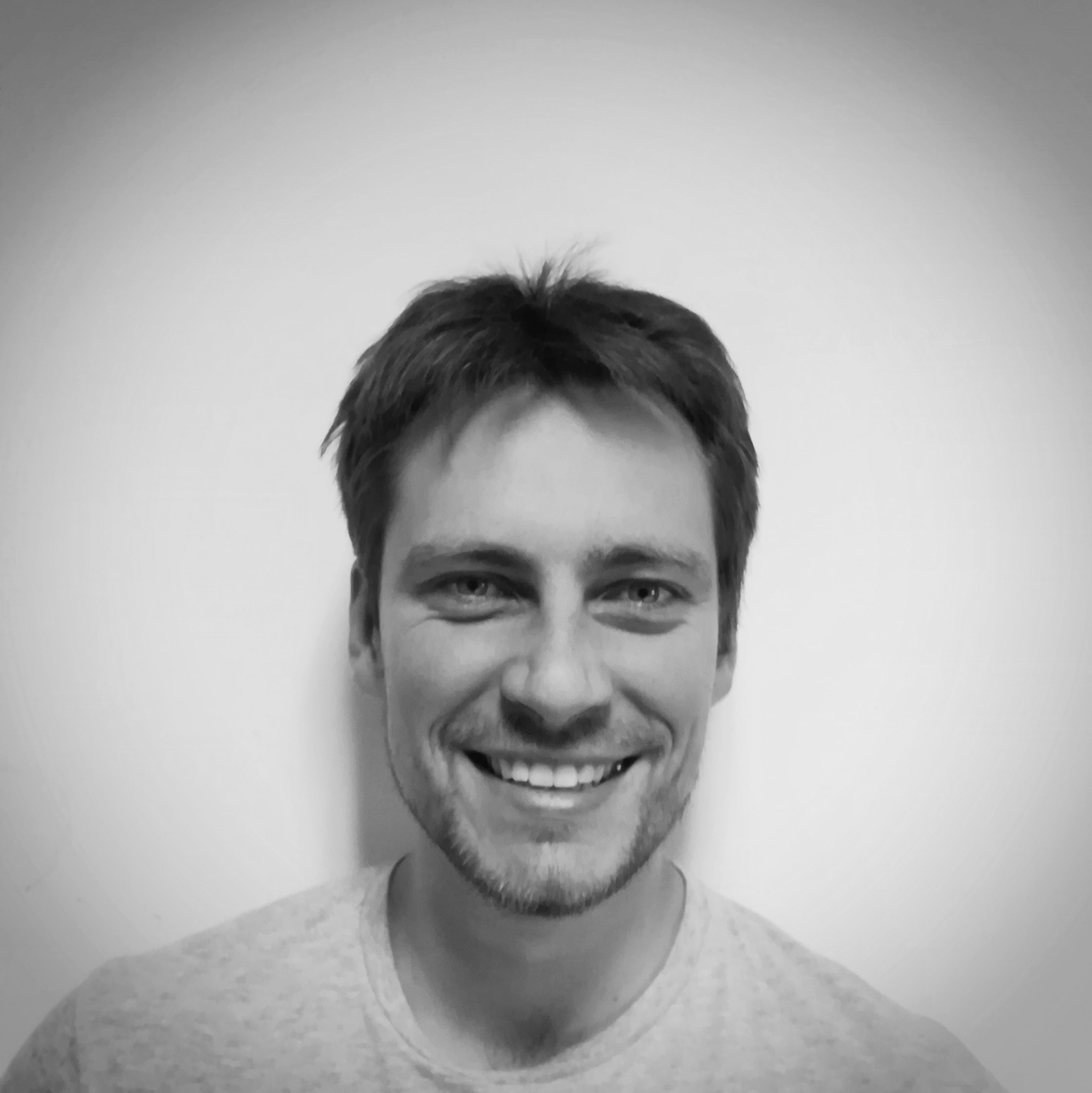}}]{Juan M. Espín López} received a B.Sc. degree in Mathematics from the University of Murcia in 2014, and a M.Sc. degree in Applied Mathematics in 2015. He received his PhD in Computer Science at the University of Murcia in 2024 and works as a Senior Machine Learning Researcher at Facephi. His research interests focus on presentation attack detection systems for documents, face and voice, speaker recognition, facial recognition, and deep learning applications in these fields.
\end{IEEEbiography}
\vspace{-0.3cm}

\begin{IEEEbiography}[{\includegraphics[width=0.9in,height=1.25in,clip,keepaspectratio]{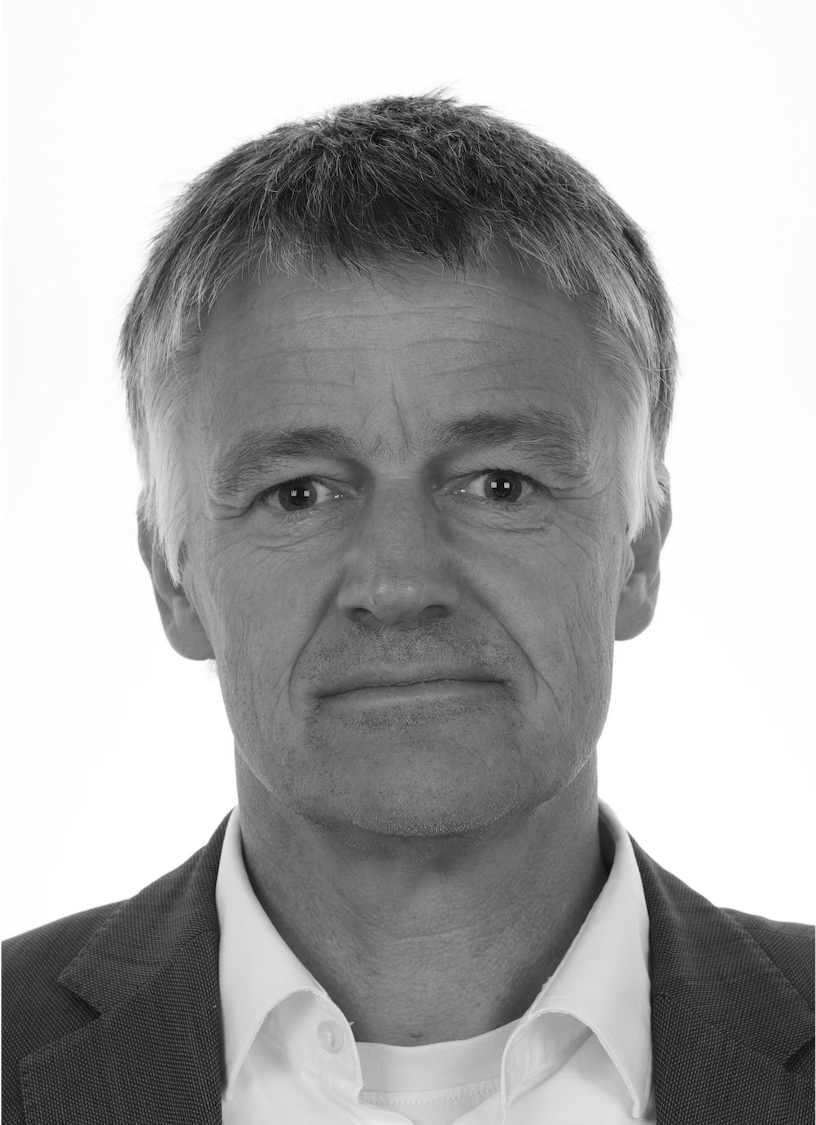}}]{Christoph Busch} is a member of the Department of Information Security and Communication Technology (IIK) at the Norwegian University of Science and Technology (NTNU), Norway. He holds a joint appointment with the computer science faculty at Hochschule Darmstadt (HDA), Germany. Further, he lectures the course Biometric Systems at Denmark’s DTU since 2007. On behalf of the German BSI he has been the coordinator for the project series BioIS, BioFace, BioFinger, BioKeyS Pilot-DB, KBEinweg and NFIQ2.0. In the European research program, he was the initiator of the Integrated Project 3D-Face, FIDELITY and iMARS. Further, he was/is partner in the projects TURBINE, BEST Network, ORIGINS, INGRESS, PIDaaS, SOTAMD, RESPECT and TReSPAsS. He is also principal investigator at the German National Research Center for Applied Cybersecurity (ATHENE). Moreover, Christoph Busch is co-founder and member of the board of the European Association for Biometrics (www.eab.org) which was established in 2011 and assembles in the meantime more than 200 institutional members. Christoph co-authored more than 700 technical papers and has been a speaker at international conferences. He is a member of the editorial board of the IET journal.
\end{IEEEbiography}

\vfill

\end{document}